\pdfoutput=1
\documentclass[11pt]{article}

\usepackage[preprint]{acl}

\usepackage{times}
\usepackage{latexsym}

\usepackage[T1]{fontenc}
\usepackage[utf8]{inputenc}

\usepackage{microtype}

\usepackage{inconsolata}

\usepackage{graphicx}

\usepackage{enumitem} 
\usepackage{amsmath, amssymb} 
\usepackage[utf8]{vietnam}
\usepackage[english]{babel}
\usepackage{multirow}
\usepackage{tabularx}
\usepackage{xcolor}
\usepackage{listings} 
\usepackage[affil-it]{authblk}
\usepackage[many]{tcolorbox}  
\usepackage{colortbl}
\usepackage{float}
\usepackage{placeins}
\usepackage{mdframed}
\usepackage{hyperref}
\usepackage{booktabs}

\tcbset{
    colback=gray!5!white,
    colframe=gray!75!black,
    boxrule=0.5pt,
    arc=4pt,
    left=4pt,
    right=4pt,
    top=4pt,
    bottom=4pt
}

\definecolor{tfidfColor}{RGB}{173, 216, 230}  % Màu xanh dương nhạt
\definecolor{qatcColor}{RGB}{144, 238, 144}   % Màu xanh lá nhạt
\title{SemViQA: A Semantic Question Answering System for Vietnamese Information Fact-Checking}

\author{
  
  \textbf{Dien X. Tran\textsuperscript{1,\protect\hyperlink{equally}{*}}}, 
  \textbf{Nam V. Nguyen\textsuperscript{1,\protect\hyperlink{equally}{*}}}, 
  \textbf{Thanh T. Tran\textsuperscript{1}},  
  \textbf{Anh T. Hoang\textsuperscript{1}}, \\
  \textbf{Tai V. Duong\textsuperscript{1}}, 
  \textbf{Di T. Le\textsuperscript{1}}, 
  \textbf{Phuc-Lu Le\textsuperscript{2}} \\
  \textsuperscript{1}Industrial University of Ho Chi Minh City, Vietnam \\ 
  \textsuperscript{2}University of Science, VNU-HCM, Vietnam \\ 
 
  \small{\textbf{Correspondence:} \href{}{lplu@fit.hcmus.edu.vn}}
}

\begin{document}
\maketitle
\def\thefootnote{*}\footnotetext{
\raisebox{\baselineskip}[0pt][0pt]{\hypertarget{equally}{}}Equal contribution.}\def\thefootnote{\arabic{footnote}}

\begin{abstract}
Recent advances in LLMs have accelerated both information generation and misinformation, especially in low-resource languages like Vietnamese, motivating robust fact-checking systems. Existing methods struggle with semantic ambiguity, homonyms, and complex linguistic structures, often trading accuracy for efficiency. We introduce SemViQA, a novel Vietnamese fact-checking framework integrating Semantic-based Evidence Retrieval (SER) and Two-step Verdict Classification (TVC). Our approach balances precision and speed, achieving state-of-the-art results with 78.97\% strict accuracy on ISE-DSC01 and 80.82\% on ViWikiFC, securing 1st place in the UIT Data Science Challenge. Additionally, SemViQA Faster improves inference speed 7× while maintaining competitive accuracy. SemViQA sets a new benchmark for Vietnamese fact verification, advancing the fight against misinformation. The source code is available at: \url{https://github.com/DAVID-NGUYEN-S16/SemViQA}.
\end{abstract}

\renewcommand{\thefootnote}{} 
\footnotetext{\raisebox{0pt}[0pt][0pt]{Preprint}}
\renewcommand{\thefootnote}{\arabic{footnote}} 
\vspace*{-0.1in}
\section{Introduction}
\vspace*{-0.1in}
\label{sec:Introduction}

\begin{figure}[h]
  \centering
  \includegraphics[width=1\linewidth]{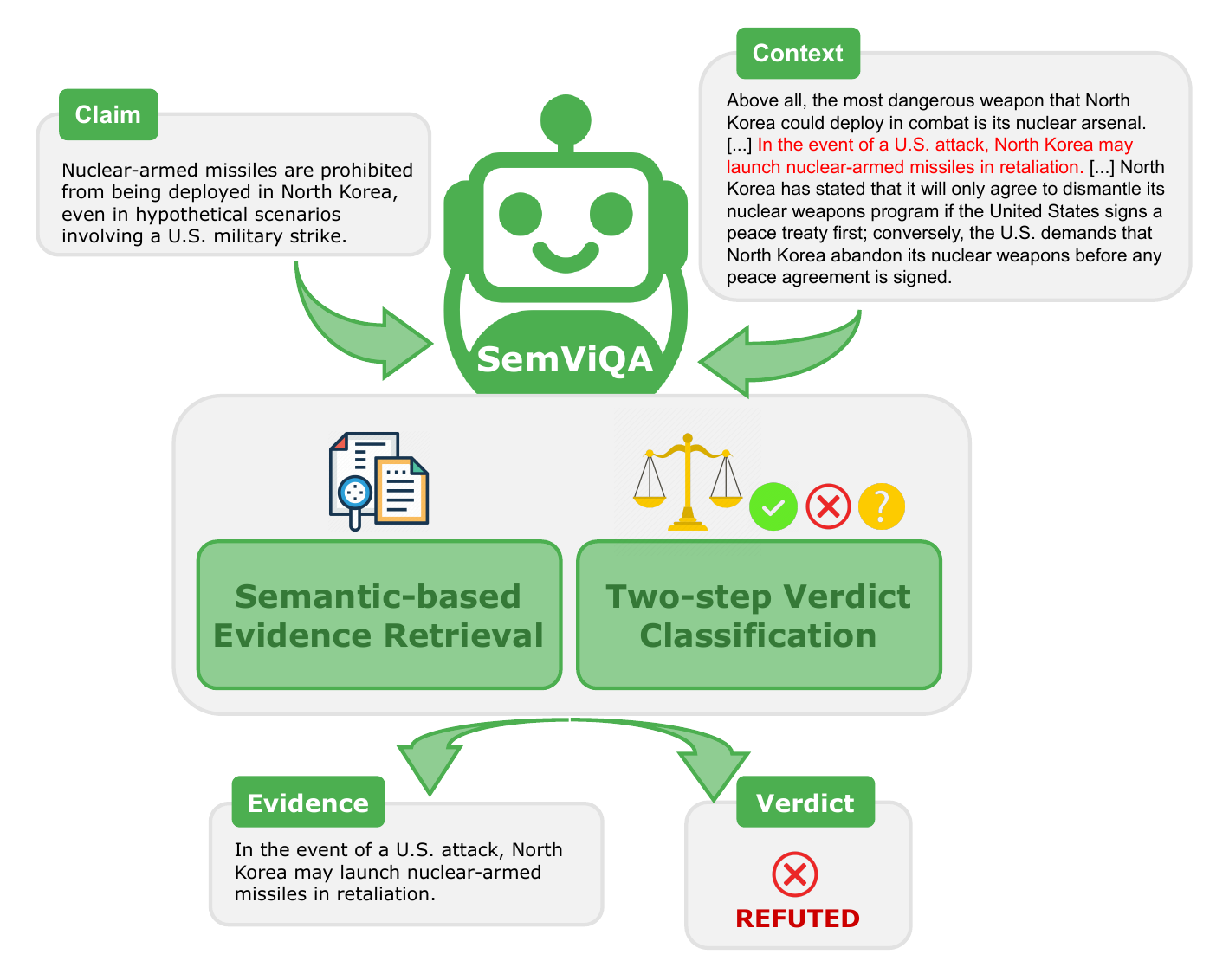}
  \caption{Overview of a Sample Information Fact-Checking Task}
  \label{fig:overview_semviqa}
\end{figure} 

The rapid advancement of large language models (LLMs), such as OpenAI’s ChatGPT, Google Gemini \cite{geminiteam2024geminifamilyhighlycapable}, Llama3.1 \cite{touvron2023llamaopenefficientfoundation}, Qwen2.5 \cite{qwen2025qwen25technicalreport}, DeepSeek V3, \cite{deepseekai2024deepseekv3technicalreport}, Phi3.5 \cite{abdin2024phi} has significantly improved information retrieval and processing across various domains. However, a major challenge with these systems is their tendency to generate factually incorrect or hallucinated content seemingly plausible information that lacks factual grounding ~\cite{ref_bert_claim_evidence}. This issue is particularly critical in domains requiring high accuracy, such as healthcare, law, and journalism, where misinformation can have serious consequences. Consequently, developing reliable fact-checking systems capable of retrieving and evaluating evidence from real-world sources has become an urgent need in Natural Language Processing (NLP).

Although fact verification has been widely studied in high-resource languages like English, applying these methods to low-resource languages such as Vietnamese remains a significant challenge. Transformer-based models, including BERT~\cite{ref_BERT} and RoBERTa~\cite{liu2019robertarobustlyoptimizedbert}, have demonstrated strong performance but their adaptation to Vietnamese is still limited. ViNSV~\cite{ref_ViNSV} employs BM25 and SBERT~\cite{ref_SBERT} for evidence retrieval but suffers from SBERT's 256-token input constraint, making it ineffective for complex, long-context claims. Graph-based reasoning methods~\cite{ref_ROLGFC} offer promising semantic inference but are often computationally expensive. Traditional retrieval methods like TF-IDF and BM25, while efficient, rely heavily on exact keyword matching, limiting their ability to capture nuanced semantics. Recent large language model (LLM) approaches~\cite{Huo_2023, schimanski-etal-2024-towards} show potential but typically require substantial computational resources, creating a trade-off between speed and accuracy.

To address these challenges, we propose \textbf{SemViQA}, a Vietnamese fact-checking framework that balances semantic accuracy and computational efficiency. As shown in Figure~\ref{fig:overview_semviqa}, SemViQA comprises three key components:

\begin{enumerate}[label=\arabic*.]
    \item \textbf{Semantic-based Evidence Retrieval (SER):} Includes a preprocessing step that efficiently handles long-token contexts by splitting them into manageable subcontexts (e.g., 400 tokens). It combines fast TF-IDF retrieval with selective Question Answering Token Classification (QATC) to strike a balance between speed and semantic accuracy.
    \item \textbf{Two-step Verdict Classification (TVC):} Employs a hierarchical classification strategy with both three-class and binary classification stages to enhance robustness and improve performance on challenging claim verification tasks.
\end{enumerate}

SemViQA achieves 78.97\% strict accuracy on ISE-DSC01~\footnote{\url{https://codalab.lisn.upsaclay.fr/competitions/15497\#results}} and 80.82\% on ViWikiFC~\cite{ref_viwikifc}, outperforming existing baselines (see Table~\ref{tab:main_results}). These results validate SemViQA's potential to enhance Vietnamese fact verification, supporting misinformation mitigation and improved transparency.

The rest of this paper is organized as follows: Section~\ref{sec:Related Works} reviews related work, Section~\ref{sec:Method} presents the methodology, Section~\ref{sec:Experiments} reports experimental results, and Section~\ref{sec:Conclusion and future works} concludes the paper with future directions.

\begin{figure*}[t]
      \includegraphics[width=1\linewidth]{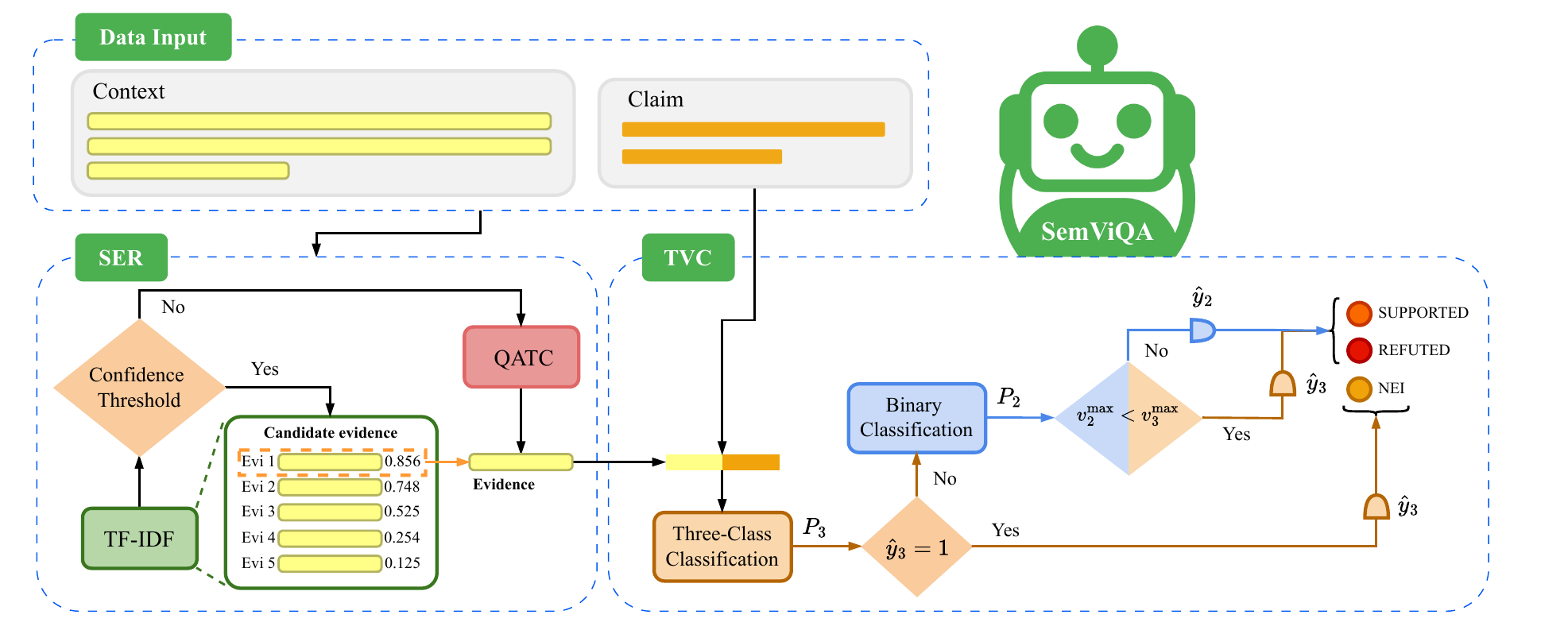} \hfill 
  \caption {\textbf{SemViQA}: A Two-Stage Method for Semantic-based Evidence Retrieval (SER) and Two-step Verdict Classification (TVC), where \( P_2 \) and \( P_3 \) represent the probabilities of the two-class and three-class classifications, respectively, and \( \hat{y}_{\text{2}} \) and \( \hat{y}_{\text{3}} \) denote their corresponding predictions.}
  \label{fig:pipeline_semviqa}
\end{figure*}
\vspace*{-0.1in}
\section{Related Works} \label{sec:Related Works}
\vspace*{-0.1in}
Advances in Natural Language Processing (NLP) have driven rapid progress in fact verification and evidence extraction. Early BiLSTM‑based models such as the Neural Semantic Matching Network (NSMN)~\cite{ref_combining}—augmented with WordNet features improved accuracy but struggled with complex sentence relations due to sequential limitations~\cite{ref_BiLSTM}. Transformer models, notably BERT~\cite{ref_BERT}, introduced bidirectional contextual encoding and achieved state‑of‑the‑art results on FEVER~\cite{ref_bert_claim_evidence, ref_gear, ref_Papelo, ref_FEVEROUS, ref_CFEVER, ref_Zero-Shot, ref_BEVERS}, yet their input‑length cap hinders long‑document fact‑checking. Graph‑based reasoning further enhances multi‑hop verification GCN variants~\cite{ref_ROLGFC, ref_Fever} and AdMIRaL’s logic‑driven retrieval~\cite{ref_NLAMDRFV} boost evidence sufficiency but at considerable computational cost. Vietnamese research is sparse: ViNSV~\cite{ref_ViNSV} pairs BM25 with SBERT~\cite{ref_SBERT} yet falters on complex reasoning due to static embeddings.

TF-IDF remains the industry standard for document retrieval due to its speed, simplicity, interpretability, and ability to handle long contexts effectively, making it suitable for rapid and scalable retrieval~\cite{reddy2018defactonlp, tfidf_paper, li2021haha, Azevedo_2022}. However, its reliance on surface-level keyword matching limits its capability to handle paraphrases, contextual nuances, and multi-hop reasoning, reducing recall accuracy on complex queries. Ensemble learning~\cite{hannichenko2023classification, wang2021ensemblelearningbasedclassification, liu2024music, ganaie2022ensemble} mitigates individual model weaknesses by aggregating diverse architectures and training signals, yielding robust gains. Building on these insights, \textbf{SemViQA} fuses fast TF‑IDF retrieval, semantic reasoning via QATC, and hierarchical classification, delivering high accuracy, low latency, and practical scalability for Vietnamese fact verification.
\vspace*{-0.1in}
\section{SemViQA - Semantic Vietnamese Question Answering}
% \vspace*{-0.1in}
\label{sec:Method}

We formulate Vietnamese fact verification as a multi-output classification task, where the input is a pair \((C, X)\), with \(C\) being a claim and \(X\) its corresponding context paragraph or document. The objective is to (i) identify the most relevant evidence sentence(s) from \(X\) and (ii) predict the veracity label of the claim as one of three categories: Supported, Refuted, or Not Enough Information (NEI). To address challenges such as long input sequences and semantic ambiguity, we propose \textbf{SemViQA}, a three-stage framework consisting of data pre-processing, semantic-based evidence retrieval, and two-step verdict classification. An overview of the architecture is shown in Figure~\ref{fig:pipeline_semviqa}, with detailed descriptions provided in the following subsections.
\vspace*{-0.1in}
\vspace*{-0.1in}
\subsection{Evidence Extraction via Question Answering with Token Classification}
\vspace*{-0.1in}
\label{subsec:QATC}
\subsubsection{Data Processing}
\label{subsec:Data Processing}

\begin{figure}[h]
  \centering
  \includegraphics[width=\columnwidth]{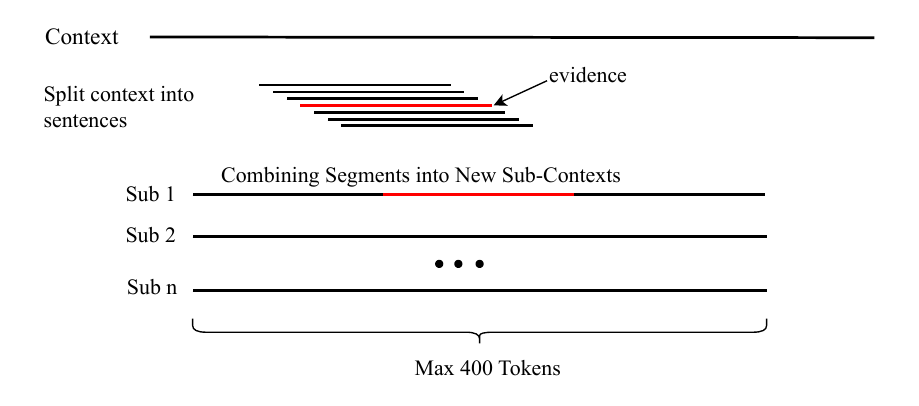}
  \caption{Long context processing solution.}
  \label{fig:long_context}
\end{figure}
To effectively support evidence retrieval and claim classification, we apply distinct preprocessing strategies to the context  based on the specific requirements of each downstream task (see Appendix~\ref{appendix:data_processing}). A key challenge arises from the considerable length of many context passages, frequently exceeding the token limits of Vietnamese BERT-based models. Figure~\ref{fig:long_context} illustrates our approach for handling long input contexts. First, the context is segmented into individual sentences. Next, sentences are sequentially aggregated into subcontexts until reaching approximately 400 tokens. Each completed subcontext is then processed by the QATC model to identify potential evidence. The next subcontext begins from the subsequent sentence, and this process continues until all sentences are processed. However, processing subcontexts sequentially can be time-consuming. Therefore, we developed SemViQA Faster, which batches and processes subcontexts in parallel, significantly accelerating the retrieval process.
\vspace*{-0.1in}
% Traditional Question Answering models predict the start and end positions of an answer span. In our framework, we extend this by incorporating a \textbf{token-level classification objective}, allowing the model to highlight relevant answer tokens with greater granularity. This dual formulation enables better supervision for evidence extraction.

% \paragraph{Start/End Span Prediction.} We concatenate the claim and context as input to a pretrained encoder and compute contextualized token embeddings. Two linear layers project the embeddings into start and end probability distributions. The span-based answer prediction is supervised using Cross-Entropy Loss:
% \begin{equation}
% \mathcal{L}_{CE} = -\sum_{i=1}^{N} t_i \log(p_i),
% \end{equation}
% where \(N\) is the number of classes (start or end positions), \(t_i\) is the true label, and \(p_i\) is the softmax probability for the \(i^{\text{th}}\) class.

% \subsubsection{Token-level Rationale Prediction}
% \label{subsubsec:rationale_token}
\subsubsection{Question Answering with Token Classification (QATC)}
\label{subsec:qatc_model}
Traditional Question Answering models typically predict the start and end positions of an answer span. In our framework, we enhance this approach by introducing a \textbf{token-level classification objective}, enabling the model to focus explicitly on tokens within evidence sentences in the context. This dual formulation provides improved supervision for evidence extraction. Drawing inspiration from rationale tagging~\cite{ju2019technicalreportconversationalquestion}, we treat token labeling as a binary classification task: tokens within evidence sentences receive a label of 1, and all other tokens receive a label of 0. In cases marked NEI (Not Enough Information), every token is labeled as 0. We employ a feed-forward classification layer on the token representations:
\begin{equation}
    p_t = \sigma\left(W_2 \cdot \text{ReLU}(W_1 h_t)\right),
\end{equation}
where \(h_t\) is the contextual representation of token \(t\), $W_1, W_2$ are learnable weights in the neural network and \(\sigma(\cdot)\) denotes the sigmoid function. The loss for this task is the Binary Cross-Entropy (BCE) loss:
% \begin{equation}
% \mathcal{L}_{RT} = -\frac{1}{T} \sum_{t=1}^{T} \left[y_t^r \log(p_t^r) + (1 - y_t^r) \log(1 - p_t^r)\right].
% \end{equation}
\begin{equation}
\mathcal{L}_{RT} = -\frac{1}{T}\sum_{t=1}^{T}\text{BCE}(y_t, p_t),
\end{equation}
% \subsubsection{Final Training Objective}
% \label{subsubsec:final_loss}

% The overall training loss integrates three components:
% \begin{equation}
% \mathcal{L} = \mathcal{L}_{CE} + \alpha \mathcal{L}_{RT},
% \end{equation}
% where \(\alpha\) is a hyperparameter balancing the rationale tagging loss.
\vspace*{-0.1in}
\subsection{Semantic-based Evidence Retrieval (SER)}
% \vspace*{-0.1in}
\label{subsec:Evidence Sentence Retrieval Method}

Accurate claim verification requires reliable evidence. To improve both efficiency and robustness, we adopt a two-stage evidence retrieval strategy combining TF-IDF with a QATC.

\textbf{Stage 1: TF-IDF-based Retrieval.}  
We segment the context \(X\) into smaller passages and pair each with the claim \(C\). Preprocessing includes noise removal and tokenization using ViTokenizer\footnote{\url{https://github.com/trungtv/pyvi}}. TF-IDF is effective for simple claims particularly refuted ones but struggles with semantically complex cases due to its reliance on keyword overlap. To enrich short segments (i.e., those with fewer than 60\% of \(C\)'s tokens), we merge them with preceding segments to improve evidence completeness. Retrieved segments are then ranked, and a confidence threshold is applied to identify easy cases (handled by TF-IDF) and hard cases (passed to QATC).

\textbf{Stage 2: QATC-based Refinement.}  
For complex cases, QATC is applied to segmented subcontexts rather than the full context due to the input length limitation of BERT models. The detailed processing approach is described in Section~\ref{subsec:Data Processing}. At this time, we consider three scenarios:  
(1) If multiple subcontexts yield conflicting answers, we collect all predicted spans and re-rank them using TF-IDF.  
(2) If a single evidence span consistently appears, it is selected directly.  
(3) If no evidence is found, fallback to TF-IDF is used.  

This hybrid approach balances speed and semantic accuracy, improving evidence selection for downstream verdict classification. Examples are provided in Appendix~\ref{appendix:tfidf_qatc_example}.
\vspace*{-0.1in}
\subsection{Two-step Verdict Classification (TVC)}
\vspace*{-0.1in}
\label{subsec:TVC}

We adopt a two-stage classification framework to enhance claim verification robustness and mitigate label imbalance, especially the overrepresentation of \textit{NEI}.

\paragraph{Stage 1: Three-Class Classification.}
Given a claim-evidence pair \((C, E)\), a BERT-based model \(f_{\text{3-class}}\) predicts a probability distribution over three labels: \textit{Supported}, \textit{Refuted}, and \textit{Not Enough Information (NEI)}:
\begin{equation}
P_3 = f_{\text{3-class}}(C, E), \quad \hat{y}_3 = \arg\max_k P_3.
\end{equation}
This step is optimized using Cross-Entropy Loss.

\paragraph{Stage 2: Binary Classification.}
If \(\hat{y}_3 \neq \text{NEI}\), we apply a refined binary classifier \(f_{\text{2-class}}\) to distinguish between \textit{Supported} and \textit{Refuted}:
\begin{equation}
P_2 = f_{\text{2-class}}(C, E), \quad \hat{y}_2 = \arg\max_k P_2.
\end{equation}
This model uses Focal Loss~\cite{ref_focal_loss} to address class imbalance.

\paragraph{Final Prediction Rule.}
The final verdict $\hat{y} \in \{1, 2, 3\}$ where  $1=\text{NEI}, \ 2 = \text{Supported}, \ 3 = \text{Refuted}$ is determined by comparing the confidence scores from both classifiers. Here, $\hat{y}$ represents the index of the predicted label, where each index corresponds to a specific class description.

\begin{equation}
\hat{y} =
\begin{cases} 
\hat{y}_{3}, & \text{if } \hat{y}_{3} = 1, \\
\hat{y}_{3}, & \text{if } v^{\text{max}}_{3} > v^{\text{max}}_{2}, \\
\hat{y}_{2}, & \text{otherwise},
\end{cases}
\end{equation}
where \(v^{\text{max}}_{3} = \max(P_{3})\), $v^{\text{max}}_{2} = \max(P_{2})$ represents the highest probability.

This hybrid strategy allows the three-class model to handle general cases, especially detecting NEI early, while the binary model specializes in distinguishing difficult SUP/REF cases.
\subsection{SemViQA Pipeline System}
We now describe the full SemViQA pipeline, as illustrated in Figure~\ref{fig:pipeline_semviqa}. First, we prepare input for TF-IDF by splitting the context paragraph $X$ into sentences, then concatenating each sentence with the claim $C$. We calculate the matching score for each sentence and select the one with the highest probability. If this score exceeds the threshold $t$, we directly use this sentence as the evidence. If the score is below $t$, we proceed to prepare input for the QATC model. We segment $X$ into subcontexts (as detailed in Section~\ref{subsec:Data Processing}). Each subcontext is sequentially processed by the QATC model. If QATC identifies zero or multiple candidate evidence spans, we collect all predicted spans and re-rank them using TF-IDF. If QATC finds exactly one candidate evidence span, we confidently select it as the final evidence. Finally, we move to the two-step verdict classification (TVC) stage. We prepare input for TVC by concatenating the claim $C$ with the final evidence. We first apply the three-class model. If it predicts NEI, the process ends. Otherwise, we use an ensemble method to combine the weights from both the three-class and binary models to make the final prediction. All coefficients used in our system are provided in the ablation study, as shown in Appendix~\ref{appendix:Ablation}.
\vspace*{-0.1in}
\section{Experiments} \label{sec:Experiments}
% \vspace*{-0.1in}
\subsection{Dataset} \label{subsec:Dataset}
We use two Vietnamese fact verification datasets: \textbf{ISE-DSC01} from the UIT Challenge 2023 and \textbf{ViWikiFC}~\cite{ref_viwikifc}, which contains over 20,000 Wikipedia-based claims, including annotated evidence for the ``nei'' label. Dataset details are provided in Appendix~\ref{appendix:data-statistics}.

\begin{table*}[t]
  \centering
  \scriptsize
  \renewcommand{\arraystretch}{1.1}
  \scalebox{0.92}{
  \begin{tabular}{ll|cccc|cccc|c}
    \hline
    \multicolumn{2}{c|}{\textbf{Method}} & \multicolumn{4}{c|}{\textbf{ViWikiFC}} & \multicolumn{4}{c|}{\textbf{ISE-DSC01}} & \textbf{Avg Strict Acc} \\
    \cline{1-2} \cline{3-10}
    ER & VC & Strict Acc & VC Acc & ER Acc & Time (s) & Strict Acc & VC Acc & ER Acc & Time (s) & \\
    \hline
    \multicolumn{11}{l}{\textbf{Traditional Baselines}} \\
    \multirow{3}{*}{TF-IDF} & InfoXLM$_{\text{large}}$ & 75.56 & 82.21 & 90.15 & 131 & 73.59 & 78.08 & 76.61 & 378 & 74.58 \\
    & XLM-R$_{\text{large}}$ & 76.47 & 82.78 & 90.15 & 134 & 75.61 & 80.50 & 78.58 & 366 & 76.04 \\
    & Ernie-M$_{\text{large}}$ & 75.56 & 81.83 & 90.15 & 144 & 78.19 & 81.69 & 80.65 & 403 & 76.88 \\
    \hline
    \multirow{3}{*}{BM25} & InfoXLM$_{\text{large}}$ & 70.44 & 79.01 & 83.50 & 130 & 72.09 & 77.37 & 75.04 & 320 & 71.27 \\
    & XLM-R$_{\text{large}}$ & 70.97 & 78.91 & 83.50 & 132 & 73.94 & 79.37 & 76.95 & 333 & 72.46 \\
    & Ernie-M$_{\text{large}}$ & 70.21 & 78.29 & 83.50 & 141 & 76.58 & 80.76 & 79.02 & 381 & 73.40 \\
    \hline
    \multirow{3}{*}{SBert} & InfoXLM$_{\text{large}}$ & 74.99 & 81.59 & 89.72 & 195 & 71.20 & 76.59 & 74.15 & 915 & 73.10 \\
    & XLM-R$_{\text{large}}$ & 75.80 & 82.35 & 89.72 & 194 & 72.85 & 78.78 & 75.89 & 835 & 74.33 \\
    & Ernie-M$_{\text{large}}$ & 75.13 & 81.44 & 89.72 & 203 & 75.46 & 79.89 & 77.91 & 920 & 75.30 \\
    \hline
    \multicolumn{11}{l}{\textbf{QA-based Approaches}} \\
    \multirow{3}{*}{ViMRC$_{\text{large}}$} & InfoXLM$_{\text{large}}$ & 77.28 & 81.97 & 92.49 & 3778 & 54.36 & 64.14 & 56.84 & 9798 & 65.82 \\
    & XLM-R$_{\text{large}}$ & 78.29 & 82.83 & 92.49 & 3824 & 53.98 & 66.70 & 57.77 & 9809 & 66.14 \\
    & Ernie-M$_{\text{large}}$ & 77.38 & 81.92 & 92.49 & 3785 & 56.62 & 62.19 & 58.91 & 9833 & 67.00 \\
    \hline
    \multirow{3}{*}{InfoXLM$_{\text{large}}$} & InfoXLM$_{\text{large}}$ & 78.14 & 82.07 & 93.45 & 4092 & 53.50 & 63.83 & 56.17 & 10057 & 65.82 \\
    & XLM-R$_{\text{large}}$ & 79.20 & 83.07 & 93.45 & 4096 & 53.32 & 66.70 & 57.25 & 10066 & 66.26 \\
    & Ernie-M$_{\text{large}}$ & 78.24 & 82.21 & 93.45 & 4102 & 56.34 & 62.36 & 58.69 & 10078 & 67.29 \\
    \hline
    \multicolumn{11}{l}{\textbf{LLMs}} \\
    \multicolumn{2}{l|}{Qwen2.5-1.5-Instruct} & 51.03 & 65.18 & 78.96 & 7665 & 59.23 & 66.68 & 65.51 & 19780 & 55.13 \\
    \multicolumn{2}{l|}{Qwen2.5-3B-Instruct} & 44.38 & 62.31 & 71.35 & 12123 & 60.87 & 66.92 & 66.10 & 31284 & 52.63 \\
    \hline
    \multirow{3}{*}{Qwen2.5-1.5-Instruct} & InfoXLM$_{\text{large}}$ & 66.14 & 76.47 & 78.96 & 7788 & 64.40 & 68.37 & 66.49 & 19970 & 65.27 \\
    & XLM-R$_{\text{large}}$ & 67.67 & 78.10 & 78.96 & 7789 & 64.66 & 69.63 & 66.72 & 19976 & 66.17 \\
    & Ernie-M$_{\text{large}}$ & 66.52 & 76.52 & 78.96 & 7794 & 65.70 & 68.37 & 67.33 & 20003 & 66.11 \\
    \hline
    \multirow{3}{*}{Qwen2.5-3B-Instruct} & InfoXLM$_{\text{large}}$ & 59.88 & 72.50 & 71.35 & 12246 & 65.72 & 69.66 & 67.51 & 31477 & 62.80 \\
    & XLM-R$_{\text{large}}$ & 60.74 & 73.08 & 71.35 & 12246 & 66.12 & 70.44 & 67.83 & 31483 & 63.43 \\
    & Ernie-M$_{\text{large}}$ & 60.02 & 72.21 & 71.35 & 12251 & 67.48 & 70.77 & 68.75 & 31512 & 63.80 \\
    \hline
    \multicolumn{11}{l}{\textbf{Ours: SER Faster + TVC}} \\
    ViMRC$_{\text{large}}$ & \multirow{2}{*}{Ernie-M$_{\text{large}}$} & \textcolor{blue}{\textbf{79.44}} & \textcolor{blue}{\textbf{82.93}} & \textcolor{blue}{\textbf{94.60}} & 410 & \textcolor{blue}{\textbf{78.32}} & \textcolor{blue}{\textbf{81.91}} & \textcolor{blue}{\textbf{80.26}} & 995 & \textcolor{blue}{\textbf{78.88}} \\
    InfoXLM$_{\text{large}}$ & & \textcolor{blue}{\textbf{79.77}} & \textcolor{blue}{\textbf{83.07}} & \textcolor{blue}{\textbf{95.03}} & 487 & \textcolor{blue}{\textbf{78.37}} & \textcolor{blue}{\textbf{81.91}} & \textcolor{blue}{\textbf{80.32}} & 925 & \textcolor{blue}{\textbf{79.07}} \\
    \hline
    \multicolumn{11}{l}{\textbf{Ours: Full SER + TVC}} \\
    \multirow{3}{*}{ViMRC$_{\text{large}}$} & InfoXLM$_{\text{large}}$ & 80.25 & 83.84 & 94.69 & 2731 & 75.13 & 79.54 & 76.87 & 5191 & 77.69 \\
    & XLM-R$_{\text{large}}$ & 80.34 & 83.64 & 94.69 & 2733 & 76.71 & 81.65 & 78.91 & 5219 & 78.53 \\
    & Ernie-M$_{\text{large}}$ & 79.53 & 82.97 & 94.69 & 2733 & \textbf{78.97} & \textbf{82.54} & \textbf{80.91} & 5225 & 79.25 \\
    \hline
    \multirow{3}{*}{InfoXLM$_{\text{large}}$} & InfoXLM$_{\text{large}}$ & 80.68 & \textbf{83.98} & \textbf{95.31} & 3860 & 75.13 & 79.60 & 76.87 & 5175 & 77.91 \\
    & XLM-R$_{\text{large}}$ & \textbf{80.82} & 83.88 & \textbf{95.31} & 3843 & 76.74 & 81.71 & 78.95 & 5200 & 78.78 \\
    & Ernie-M$_{\text{large}}$ & 80.06 & 83.17 & \textbf{95.31} & 3891 & \textbf{78.97} & 82.49 & \textbf{80.91} & 5297 & \textbf{79.52} \\
    \hline
  \end{tabular}
  }
  \caption{Performance comparison on the ViWikiFC test set and the ISE-DSC01 private-test dataset. The results highlight differences among models based on several criteria: Strict Accuracy (Strict Acc), Veracity Classification Accuracy (VC Acc), and Evidence Retrieval Accuracy (ER Acc). Time represents the total inference time required to generate the complete results.}
  \label{tab:main_results}
\end{table*}

\subsection{Experimental Setup} 
\label{subsec:Experimental Setup}

We conducted extensive experiments on NVIDIA A100 GPUs, fine-tuning key hyperparameters while keeping consistent settings across runs. The final configuration, selected via rigorous validation, improved both accuracy and strict accuracy on ISE-DSC01 and ViWikiFC. Full details are provided in Appendix~\ref{appendix:hyperparameter_llm}. For fair evaluation, all methods were tested on a Kaggle instance with an NVIDIA T4 GPU.

The large language model was fine-tuned in a distributed A100 setup using a structured prompt-based reformulation. Raw data were converted into prompt format to align with LLM training objectives and maximize task-specific performance. Training setup, prompt design, and preprocessing are also detailed in Appendix~\ref{appendix:hyperparameter_llm}.
\vspace*{-0.1in}
\subsection{Main Results} 
\vspace*{-0.1in}
\label{subsec:Experimental Results}

The results in Table \ref{tab:main_results} demonstrate that SemViQA outperforms previous methods in Vietnamese fact-checking tasks. Specifically, our model achieves the highest Strict Accuracy, reaching 80.82\% on ViWikiFC and 78.97\% on ISE-DSC01, establishing a new benchmark for automated fact-checking systems in Vietnamese language.
\subsubsection{Performance Comparison}

\paragraph{a) Handling Long Token Sequences in Fact-Checking}
A major limitation of conventional Question Answering (QA) models in fact verification is their inability to process long-context claims due to the 512-token input limit of transformer-based models such as ViMRC\(_{large}\)\footnote{\url{https://huggingface.co/nguyenvulebinh/vi-mrc-large}}, InfoXLM$_{large}$~~\cite{ref_infoXLM}, XLM-R$_{large}$~~\cite{ref_XLM_R}, and Ernie-M$_{large}$~\cite{ouyang-etal-2021-ernie}. Real-world datasets like ISE-DSC01 often contain contexts exceeding 4800 tokens, severely degrading QA-based performance by limiting access to full evidence. To overcome this, SemViQA employs an efficient retrieval-based strategy (see Section~\ref{subsec:Data Processing}) that handles long-token sequences effectively. On ISE-DSC01, SemViQA outperforms traditional QA models by fully leveraging extended contexts, confirming that the long-token constraint is a critical bottleneck. Conversely, on ViWikiFC, where contexts average around 512 tokens, QA models perform competitively. Yet, even in this setting, integrating our Semantic-based Evidence Retrieval (SER) yields a 1.86\% improvement in evidence retrieval accuracy, demonstrating the versatility and efficiency of our approach. These findings emphasize that long-token limitations significantly hinder fact verification, and SemViQA successfully mitigates this issue while enhancing QA models across varying dataset conditions.

% \paragraph{b) Performance and Inference Time Optimization}
% One of the key advancements of SemViQA over other methods is its significant reduction in inference time without compromising accuracy. Key observations include:

% \begin{itemize}
%     \item The average inference time of SemViQA on ISE-DSC01 is 5200s, whereas large LLM-based models such as Qwen2.5-3B-Instruct~\cite{qwen2024qwen25} require over 31,000s, making SemViQA at least \textbf{6 times faster} than these large language model approaches.
%     \item Compared to ViMRC\(_{large}\) (9800s on ISE-DSC01), SemViQA reduces inference time by nearly 50\% while maintaining higher performance in both Strict Accuracy and Veracity Classification Accuracy.
%     \item Methods based on SBERT~\cite{reimers-gurevych-2019-sentence} or BM25 have lower inference times but fail to maintain high accuracy, particularly when dealing with multi-step reasoning tasks. SemViQA achieves a better balance between classification performance and processing speed, ensuring practical deployment in real-world environments.
% \end{itemize}
\vspace{-0.1cm}
\begin{figure}[t]
  \centering
  \includegraphics[width=1\linewidth]{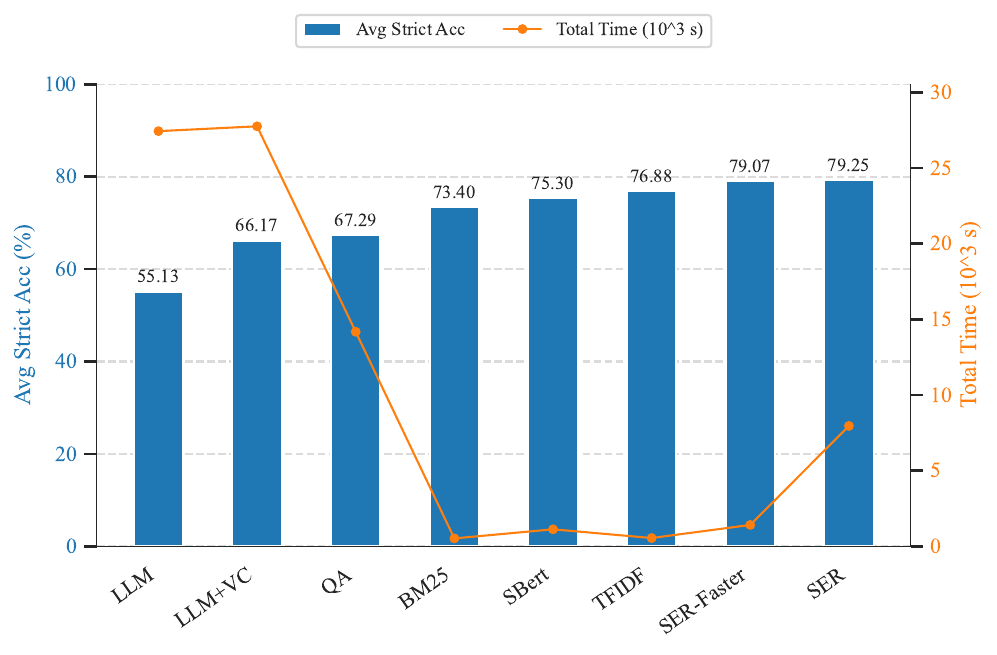}
  \caption{Comparison of methods in terms of peak performance and total inference time across datasets. Each retrieval approach is evaluated by its best score, while overall efficiency is reflected through cumulative inference time. See Table~\ref{tab:main_results} for details.}

  \label{fig:accuracy_vs_time}
\end{figure}
\vspace{-0.1cm}
% \textbf{SemViQA Faster}: We introduce an optimized version of SemViQA that achieves up to 7× faster inference while maintaining high accuracy. As shown in Figure ~\ref{fig:accuracy_vs_time}, the inference speed of SER-Faster closely matches that of traditional methods while still outperforming competing solutions in both efficiency and accuracy. To achieve this speed improvement, we process all subcontexts within a sample as a batch, enabling batch inference (see Section~\ref{subsec:Data Processing} for details on subcontexts). While this approach results in a slight performance drop compared to the standard SemViQA model, the trade-off is minimal considering the significant reduction in processing time. This efficiency enhancement makes SemViQA Faster highly suitable for real-world deployment, allowing seamless integration into large-scale fact-checking applications.
\paragraph{b) Performance and Inference Time Optimization}
SemViQA significantly reduces inference time while maintaining high accuracy, making it highly practical for real-world applications. Key highlights include:

\begin{itemize}
\item On ISE-DSC01, SemViQA averages 5200s per run, over \textbf{6 times faster} than large LLM-based models like Qwen2.5-3B-Instruct~\cite{qwen2024qwen25}, which require over 31,000s.
\item Compared to ViMRC$_{large}$ (9800s), SemViQA halves inference time while achieving superior Strict Accuracy and Veracity Classification Accuracy.
\item Although BM25 and SBERT~\cite{reimers-gurevych-2019-sentence} are faster, they struggle with complex, multi-step reasoning where SemViQA maintains a strong balance between speed and accuracy.
\end{itemize}

\textbf{SemViQA Faster}: We further introduce SemViQA Faster, which accelerates inference by up to \textbf{7$\times$} through batch processing of subcontexts (see Section~\ref{subsec:Data Processing}). As shown in Figure~\ref{fig:accuracy_vs_time}, this variant achieves inference speeds comparable to traditional methods while retaining competitive accuracy. The minor performance trade-off is acceptable given the substantial time savings, making SemViQA Faster ideal for scalable, real-world fact-checking systems.

\subsubsection{Comparison with other results in the competition}

\begin{table}[h]
  \centering
  \resizebox{0.85\columnwidth}{!}{%
    \begin{tabular}{lccc} 
      \hline
      \textbf{Methods} & \textbf{Strict Acc} & \textbf{VC Acc} & \textbf{ER Acc} \\  
      \hline
      \textbf{SemViQA} &  \textbf{78.97} & 82.54  & \textbf{80.91} \\ 
      DS@UIT Dynasty\footnote{\url{https://github.com/minhquan6203/Verdict-Classification-for-Fact-Checking-at-DSC-2023}} &  78.05 & \textbf{84.76}  & 80.13 \\
      URA\_FNU\footnote{\url{https://github.com/virrosluo/URA_UIT_Data_Science_Challenge}} &  77.87 & 83.71  & 79.96 \\ 
      ViNSV \cite{ref_ViNSV} &  76.33 & 81.67  & 78.11 \\
      \cite{Tran_2024} &  75.11 &  82.30  & 76.82 \\
      \hline
    \end{tabular}%
  }
  \caption{Private leaderboard comparison of top systems in the ISE-DSC01 competition.}
  \label{tab3:top5_competion}
\end{table}

The results presented in Table \ref{tab3:top5_competion} indicate that our SemViQA approach outperforms other competing teams, achieving the highest Strict Accuracy and demonstrating exceptional effectiveness in information processing and verification. This achievement highlights SemViQA's capability to deliver significantly more accurate and reliable results.

\section{Conclusion and Future Works} \label{sec:Conclusion and future works}

We introduced SemViQA, a Vietnamese fact-checking framework that integrates Semantic-based Evidence Retrieval (SER) and Two-step Verdict Classification (TVC) to enhance claim verification. Our approach outperforms existing methods, including LLMs, TF-IDF, BM25, SBERT, and QA-based models, particularly in handling long-token sequences and complex reasoning tasks. Extensive experiments demonstrated SemViQA's state-of-the-art performance on ISE-DSC01 and ViWikiFC. Additionally, the SemViQA Faster variant accelerates inference by up to 7×, improving its practicality for real-world applications. By addressing key challenges such as semantic ambiguity and multi-step reasoning, SemViQA lays the groundwork for advancing Vietnamese NLP, with potential applications in misinformation detection and low-resource language fact-checking.
\newpage
\section*{Limitations} 
% \newline
While SemViQA demonstrates strong performance in Vietnamese fact verification, several limitations remain. First, our reliance on TF-IDF for initial evidence retrieval, while efficient, limits the model’s ability to capture deep semantic relationships and retrieve implicit evidence. To mitigate this, we employ a threshold-based mechanism to identify hard samples and process them with a more advanced retrieval model. However, this approach relies on manually defined thresholds, which may not generalize well across different datasets, underscoring the need for adaptive and data-driven retrieval strategies in future work. Second, our Two-step Verdict Classification (TVC) framework improves claim verification accuracy but requires multiple classification stages, increasing inference time compared to single-step approaches. This additional computational cost is particularly significant in three-class classification tasks, where optimizing model efficiency without compromising accuracy remains a key challenge. Future work should focus on refining retrieval mechanisms and classification strategies to enhance efficiency and robustness, ensuring broader applicability of SemViQA in real-world fact verification scenarios.

% ------------------------------------------
\bibliography{custom}

\begin{thebibliography}{40}
\providecommand{\natexlab}[1]{#1}

\bibitem[{Abdin et~al.(2024)Abdin, Jacobs, Awan, Aneja, Awadallah, Awadalla, Bach, Bahree, Bakhtiari, Behl et~al.}]{abdin2024phi}
Marah Abdin, Sam~Ade Jacobs, Ammar~Ahmad Awan, Jyoti Aneja, Ahmed Awadallah, Hany Awadalla, Nguyen Bach, Amit Bahree, Arash Bakhtiari, Harkirat Behl, et~al. 2024.
\newblock Phi-3 technical report: A highly capable language model locally on your phone.
\newblock \emph{arXiv preprint arXiv:2404.14219}.

\bibitem[{Aly et~al.(2021)Aly, Guo, Schlichtkrull, Thorne, Vlachos, Christodoulopoulos, Cocarascu, and Mittal}]{ref_FEVEROUS}
Rami Aly, Zhijiang Guo, Michael~Sejr Schlichtkrull, James Thorne, Andreas Vlachos, Christos Christodoulopoulos, Oana Cocarascu, and Arpit Mittal. 2021.
\newblock \href {https://doi.org/10.18653/v1/2021.fever-1.1} {The fact extraction and {VER}ification over unstructured and structured information ({FEVEROUS}) shared task}.
\newblock In \emph{Proceedings of the Fourth Workshop on Fact Extraction and VERification (FEVER)}, pages 1--13, Dominican Republic. Association for Computational Linguistics.

\bibitem[{Aly and Vlachos(2022)}]{ref_NLAMDRFV}
Rami Aly and Andreas Vlachos. 2022.
\newblock \href {https://doi.org/10.18653/v1/2022.emnlp-main.411} {Natural logic-guided autoregressive multi-hop document retrieval for fact verification}.
\newblock In \emph{Proceedings of the 2022 Conference on Empirical Methods in Natural Language Processing}, pages 6123--6135, Abu Dhabi, United Arab Emirates. Association for Computational Linguistics.

\bibitem[{Azevedo et~al.(2022)Azevedo, Rocha, Esteves, and Cardoso}]{Azevedo_2022}
Pedro Azevedo, Gil Rocha, Diego Esteves, and Henrique~Lopes Cardoso. 2022.
\newblock \href {https://doi.org/10.1145/3486622.3493930} {Towards better evidence extraction methods for fact-checking systems}.
\newblock In \emph{IEEE/WIC/ACM International Conference on Web Intelligence and Intelligent Agent Technology}, WI-IAT '21, page 277–284, New York, NY, USA. Association for Computing Machinery.

\bibitem[{Chi et~al.(2021)Chi, Dong, Wei, Yang, Singhal, Wang, Song, Mao, Huang, and Zhou}]{ref_infoXLM}
Zewen Chi, Li~Dong, Furu Wei, Nan Yang, Saksham Singhal, Wenhui Wang, Xia Song, Xian-Ling Mao, Heyan Huang, and Ming Zhou. 2021.
\newblock \href {https://doi.org/10.18653/v1/2021.naacl-main.280} {{I}nfo{XLM}: An information-theoretic framework for cross-lingual language model pre-training}.
\newblock In \emph{Proceedings of the 2021 Conference of the North American Chapter of the Association for Computational Linguistics: Human Language Technologies}, pages 3576--3588, Online. Association for Computational Linguistics.

\bibitem[{Conneau et~al.(2020)Conneau, Khandelwal, Goyal, Chaudhary, Wenzek, Guzm{\'a}n, Grave, Ott, Zettlemoyer, and Stoyanov}]{ref_XLM_R}
Alexis Conneau, Kartikay Khandelwal, Naman Goyal, Vishrav Chaudhary, Guillaume Wenzek, Francisco Guzm{\'a}n, Edouard Grave, Myle Ott, Luke Zettlemoyer, and Veselin Stoyanov. 2020.
\newblock \href {https://doi.org/10.18653/v1/2020.acl-main.747} {Unsupervised cross-lingual representation learning at scale}.
\newblock In \emph{Proceedings of the 58th Annual Meeting of the Association for Computational Linguistics}, pages 8440--8451, Online. Association for Computational Linguistics.

\bibitem[{DeepSeek-AI et~al.(2024)DeepSeek-AI, Liu, Feng, Xue, Wang, Wu, Lu, Zhao, Deng, Zhang, Ruan, Dai, Guo, Yang, Chen, Ji, Li, Lin, Dai, Luo, Hao, Chen, Li, Zhang, Bao, Xu, Wang, Zhang, Ding, Xin, Gao, Li, Qu, Cai, Liang, Guo, Ni, Li, Wang, Chen, Chen, Yuan, Qiu, Li, Song, Dong, Hu, Gao, Guan, Huang, Yu, Wang, Zhang, Xu, Xia, Zhao, Wang, Zhang, Li, Wang, Zhang, Zhang, Tang, Li, Tian, Huang, Wang, Zhang, Wang, Zhu, Chen, Du, Chen, Jin, Ge, Zhang, Pan, Wang, Xu, Zhang, Chen, Li, Lu, Zhou, Chen, Wu, Ye, Ye, Ma, Wang, Zhou, Yu, Zhou, Pan, Wang, Yun, Pei, Sun, Xiao, Zeng, Zhao, An, Liu, Liang, Gao, Yu, Zhang, Li, Jin, Wang, Bi, Liu, Wang, Shen, Chen, Zhang, Chen, Nie, Sun, Wang, Cheng, Liu, Xie, Liu, Yu, Song, Shan, Zhou, Yang, Li, Su, Lin, Li, Wang, Wei, Zhu, Zhang, Xu, Xu, Huang, Li, Zhao, Sun, Li, Wang, Yu, Zheng, Zhang, Shi, Xiong, He, Tang, Piao, Wang, Tan, Ma, Liu, Guo, Wu, Ou, Zhu, Wang, Gong, Zou, He, Zha, Xiong, Ma, Yan, Luo, You, Liu, Zhou, Wu, Ren, Ren, Sha, Fu, Xu, Huang, Zhang, Xie, Zhang, Hao,
  Gou, Ma, Yan, Shao, Xu, Wu, Zhang, Li, Gu, Zhu, Liu, Li, Xie, Song, Gao, and Pan}]{deepseekai2024deepseekv3technicalreport}
DeepSeek-AI, Aixin Liu, Bei Feng, Bing Xue, Bingxuan Wang, Bochao Wu, Chengda Lu, Chenggang Zhao, Chengqi Deng, Chenyu Zhang, Chong Ruan, Damai Dai, Daya Guo, Dejian Yang, Deli Chen, Dongjie Ji, Erhang Li, Fangyun Lin, Fucong Dai, Fuli Luo, Guangbo Hao, Guanting Chen, Guowei Li, H.~Zhang, Han Bao, Hanwei Xu, Haocheng Wang, Haowei Zhang, Honghui Ding, Huajian Xin, Huazuo Gao, Hui Li, Hui Qu, J.~L. Cai, Jian Liang, Jianzhong Guo, Jiaqi Ni, Jiashi Li, Jiawei Wang, Jin Chen, Jingchang Chen, Jingyang Yuan, Junjie Qiu, Junlong Li, Junxiao Song, Kai Dong, Kai Hu, Kaige Gao, Kang Guan, Kexin Huang, Kuai Yu, Lean Wang, Lecong Zhang, Lei Xu, Leyi Xia, Liang Zhao, Litong Wang, Liyue Zhang, Meng Li, Miaojun Wang, Mingchuan Zhang, Minghua Zhang, Minghui Tang, Mingming Li, Ning Tian, Panpan Huang, Peiyi Wang, Peng Zhang, Qiancheng Wang, Qihao Zhu, Qinyu Chen, Qiushi Du, R.~J. Chen, R.~L. Jin, Ruiqi Ge, Ruisong Zhang, Ruizhe Pan, Runji Wang, Runxin Xu, Ruoyu Zhang, Ruyi Chen, S.~S. Li, Shanghao Lu, Shangyan Zhou, Shanhuang
  Chen, Shaoqing Wu, Shengfeng Ye, Shengfeng Ye, Shirong Ma, Shiyu Wang, Shuang Zhou, Shuiping Yu, Shunfeng Zhou, Shuting Pan, T.~Wang, Tao Yun, Tian Pei, Tianyu Sun, W.~L. Xiao, Wangding Zeng, Wanjia Zhao, Wei An, Wen Liu, Wenfeng Liang, Wenjun Gao, Wenqin Yu, Wentao Zhang, X.~Q. Li, Xiangyue Jin, Xianzu Wang, Xiao Bi, Xiaodong Liu, Xiaohan Wang, Xiaojin Shen, Xiaokang Chen, Xiaokang Zhang, Xiaosha Chen, Xiaotao Nie, Xiaowen Sun, Xiaoxiang Wang, Xin Cheng, Xin Liu, Xin Xie, Xingchao Liu, Xingkai Yu, Xinnan Song, Xinxia Shan, Xinyi Zhou, Xinyu Yang, Xinyuan Li, Xuecheng Su, Xuheng Lin, Y.~K. Li, Y.~Q. Wang, Y.~X. Wei, Y.~X. Zhu, Yang Zhang, Yanhong Xu, Yanhong Xu, Yanping Huang, Yao Li, Yao Zhao, Yaofeng Sun, Yaohui Li, Yaohui Wang, Yi~Yu, Yi~Zheng, Yichao Zhang, Yifan Shi, Yiliang Xiong, Ying He, Ying Tang, Yishi Piao, Yisong Wang, Yixuan Tan, Yiyang Ma, Yiyuan Liu, Yongqiang Guo, Yu~Wu, Yuan Ou, Yuchen Zhu, Yuduan Wang, Yue Gong, Yuheng Zou, Yujia He, Yukun Zha, Yunfan Xiong, Yunxian Ma, Yuting Yan, Yuxiang
  Luo, Yuxiang You, Yuxuan Liu, Yuyang Zhou, Z.~F. Wu, Z.~Z. Ren, Zehui Ren, Zhangli Sha, Zhe Fu, Zhean Xu, Zhen Huang, Zhen Zhang, Zhenda Xie, Zhengyan Zhang, Zhewen Hao, Zhibin Gou, Zhicheng Ma, Zhigang Yan, Zhihong Shao, Zhipeng Xu, Zhiyu Wu, Zhongyu Zhang, Zhuoshu Li, Zihui Gu, Zijia Zhu, Zijun Liu, Zilin Li, Ziwei Xie, Ziyang Song, Ziyi Gao, and Zizheng Pan. 2024.
\newblock \href {https://arxiv.org/abs/2412.19437} {Deepseek-v3 technical report}.
\newblock \emph{Preprint}, arXiv:2412.19437.

\bibitem[{DeHaven and Scott(2023)}]{ref_BEVERS}
Mitchell DeHaven and Stephen Scott. 2023.
\newblock \href {https://doi.org/10.18653/v1/2023.fever-1.6} {{BEVERS}: A general, simple, and performant framework for automatic fact verification}.
\newblock In \emph{Proceedings of the Sixth Fact Extraction and VERification Workshop (FEVER)}, pages 58--65, Dubrovnik, Croatia. Association for Computational Linguistics.

\bibitem[{Devlin et~al.(2019)Devlin, Chang, Lee, and Toutanova}]{ref_BERT}
Jacob Devlin, Ming-Wei Chang, Kenton Lee, and Kristina Toutanova. 2019.
\newblock \href {https://doi.org/10.18653/v1/N19-1423} {{BERT}: Pre-training of deep bidirectional transformers for language understanding}.
\newblock In \emph{Proceedings of the 2019 Conference of the North {A}merican Chapter of the Association for Computational Linguistics: Human Language Technologies, Volume 1 (Long and Short Papers)}, pages 4171--4186, Minneapolis, Minnesota. Association for Computational Linguistics.

\bibitem[{Ganaie et~al.(2022)Ganaie, Hu, Malik, Tanveer, and Suganthan}]{ganaie2022ensemble}
Mudasir~A Ganaie, Minghui Hu, Ashwani~Kumar Malik, Muhammad Tanveer, and Ponnuthurai~N Suganthan. 2022.
\newblock Ensemble deep learning: A review.
\newblock \emph{Engineering Applications of Artificial Intelligence}, 115:105151.

\bibitem[{Graves and Schmidhuber(2005)}]{ref_BiLSTM}
Alex Graves and Jürgen Schmidhuber. 2005.
\newblock \href {https://doi.org/10.1016/j.neunet.2005.06.042} {Framewise phoneme classification with bidirectional lstm and other neural network architectures}.
\newblock \emph{Neural networks : the official journal of the International Neural Network Society}, 18:602--10.

\bibitem[{Hannichenko et~al.(2023)Hannichenko, Bidyuk, Kalinina, and Zhebko}]{hannichenko2023classification}
Tetyana Hannichenko, Peter Bidyuk, Irina Kalinina, and Oleksandr Zhebko. 2023.
\newblock Classification system based on ensemble methods for solving machine learning tasks.

\bibitem[{Huo et~al.(2023)Huo, Arabzadeh, and Clarke}]{Huo_2023}
Siqing Huo, Negar Arabzadeh, and Charles Clarke. 2023.
\newblock \href {https://doi.org/10.1145/3624918.3625336} {Retrieving supporting evidence for generative question answering}.
\newblock In \emph{Proceedings of the Annual International ACM SIGIR Conference on Research and Development in Information Retrieval in the Asia Pacific Region}, SIGIR-AP ’23, page 11–20. ACM.

\bibitem[{Ju et~al.(2019)Ju, Zhao, Chen, Zheng, Yang, and Liu}]{ju2019technicalreportconversationalquestion}
Ying Ju, Fubang Zhao, Shijie Chen, Bowen Zheng, Xuefeng Yang, and Yunfeng Liu. 2019.
\newblock \href {https://arxiv.org/abs/1909.10772} {Technical report on conversational question answering}.
\newblock \emph{Preprint}, arXiv:1909.10772.

\bibitem[{Le et~al.(2024)Le, To, Nguyen, and Nguyen}]{ref_viwikifc}
Hung~Tuan Le, Long~Truong To, Manh~Trong Nguyen, and Kiet~Van Nguyen. 2024.
\newblock \href {https://arxiv.org/abs/2405.07615} {Viwikifc: Fact-checking for vietnamese wikipedia-based textual knowledge source}.
\newblock \emph{Preprint}, arXiv:2405.07615.

\bibitem[{Li(2021)}]{li2021haha}
Kun Li. 2021.
\newblock Haha at fakedes 2021: A fake news detection method based on tf-idf and ensemble machine learning.
\newblock In \emph{IberLEF@ SEPLN}, pages 630--638.

\bibitem[{Lin et~al.(2018)Lin, Goyal, Girshick, He, and Dollár}]{ref_focal_loss}
Tsung-Yi Lin, Priya Goyal, Ross Girshick, Kaiming He, and Piotr Dollár. 2018.
\newblock \href {https://arxiv.org/abs/1708.02002} {Focal loss for dense object detection}.
\newblock \emph{Preprint}, arXiv:1708.02002.

\bibitem[{Lin et~al.(2024)Lin, Lin, Yeh, Li, Hu, Hsu, Lee, and Kao}]{ref_CFEVER}
Ying-Jia Lin, Chun-Yi Lin, Chia-Jen Yeh, Yi-Ting Li, Yun-Yu Hu, Chih-Hao Hsu, Mei-Feng Lee, and Hung-Yu Kao. 2024.
\newblock \href {https://arxiv.org/abs/2402.13025} {Cfever: A chinese fact extraction and verification dataset}.
\newblock \emph{Preprint}, arXiv:2402.13025.

\bibitem[{Liu et~al.(2024)Liu, Dasgupta, and He}]{liu2024music}
Yichen Liu, Abhijit Dasgupta, and Qiwei He. 2024.
\newblock \href {https://arxiv.org/abs/2412.15602} {Music genre classification: Ensemble learning with subcomponents-level attention}.
\newblock \emph{Preprint}, arXiv:2412.15602.

\bibitem[{Liu et~al.(2019)Liu, Ott, Goyal, Du, Joshi, Chen, Levy, Lewis, Zettlemoyer, and Stoyanov}]{liu2019robertarobustlyoptimizedbert}
Yinhan Liu, Myle Ott, Naman Goyal, Jingfei Du, Mandar Joshi, Danqi Chen, Omer Levy, Mike Lewis, Luke Zettlemoyer, and Veselin Stoyanov. 2019.
\newblock \href {https://arxiv.org/abs/1907.11692} {Roberta: A robustly optimized bert pretraining approach}.
\newblock \emph{Preprint}, arXiv:1907.11692.

\bibitem[{Malon(2018)}]{ref_Papelo}
Christopher Malon. 2018.
\newblock \href {https://doi.org/10.18653/v1/W18-5517} {Team papelo: Transformer networks at {FEVER}}.
\newblock In \emph{Proceedings of the First Workshop on Fact Extraction and {VER}ification ({FEVER})}, pages 109--113, Brussels, Belgium. Association for Computational Linguistics.

\bibitem[{Nie et~al.(2018)Nie, Chen, and Bansal}]{ref_combining}
Yixin Nie, Haonan Chen, and Mohit Bansal. 2018.
\newblock \href {https://arxiv.org/abs/1811.07039} {Combining fact extraction and verification with neural semantic matching networks}.
\newblock \emph{Preprint}, arXiv:1811.07039.

\bibitem[{Ouyang et~al.(2021)Ouyang, Wang, Pang, Sun, Tian, Wu, and Wang}]{ouyang-etal-2021-ernie}
Xuan Ouyang, Shuohuan Wang, Chao Pang, Yu~Sun, Hao Tian, Hua Wu, and Haifeng Wang. 2021.
\newblock \href {https://doi.org/10.18653/v1/2021.emnlp-main.3} {{ERNIE}-{M}: Enhanced multilingual representation by aligning cross-lingual semantics with monolingual corpora}.
\newblock In \emph{Proceedings of the 2021 Conference on Empirical Methods in Natural Language Processing}, pages 27--38, Online and Punta Cana, Dominican Republic. Association for Computational Linguistics.

\bibitem[{Qaiser and Ali(2018)}]{tfidf_paper}
Shahzad Qaiser and Ramsha Ali. 2018.
\newblock \href {https://doi.org/10.5120/ijca2018917395} {Text mining: Use of tf-idf to examine the relevance of words to documents}.
\newblock \emph{International Journal of Computer Applications}, 181.

\bibitem[{Qwen et~al.(2025)Qwen, :, Yang, Yang, Zhang, Hui, Zheng, Yu, Li, Liu, Huang, Wei, Lin, Yang, Tu, Zhang, Yang, Yang, Zhou, Lin, Dang, Lu, Bao, Yang, Yu, Li, Xue, Zhang, Zhu, Men, Lin, Li, Tang, Xia, Ren, Ren, Fan, Su, Zhang, Wan, Liu, Cui, Zhang, and Qiu}]{qwen2025qwen25technicalreport}
Qwen, :, An~Yang, Baosong Yang, Beichen Zhang, Binyuan Hui, Bo~Zheng, Bowen Yu, Chengyuan Li, Dayiheng Liu, Fei Huang, Haoran Wei, Huan Lin, Jian Yang, Jianhong Tu, Jianwei Zhang, Jianxin Yang, Jiaxi Yang, Jingren Zhou, Junyang Lin, Kai Dang, Keming Lu, Keqin Bao, Kexin Yang, Le~Yu, Mei Li, Mingfeng Xue, Pei Zhang, Qin Zhu, Rui Men, Runji Lin, Tianhao Li, Tianyi Tang, Tingyu Xia, Xingzhang Ren, Xuancheng Ren, Yang Fan, Yang Su, Yichang Zhang, Yu~Wan, Yuqiong Liu, Zeyu Cui, Zhenru Zhang, and Zihan Qiu. 2025.
\newblock \href {https://arxiv.org/abs/2412.15115} {Qwen2.5 technical report}.
\newblock \emph{Preprint}, arXiv:2412.15115.

\bibitem[{Qwen et~al.(2024)Qwen, :, Yang, Yang, Zhang, Hui, Zheng, Yu, Li, Liu, Huang, Wei, Lin, Yang, Tu, Zhang, Yang, Yang, Zhou, Lin, Dang, Lu, Bao, Yang, Yu, Li, Xue, Zhang, Zhu, Men, Lin, Li, Tang, Xia, Ren, Ren, Fan, Su, Zhang, Wan, Liu, Cui, Zhang, and Qiu}]{qwen2024qwen25}
Qwen, :, An~Yang, Baosong Yang, Beichen Zhang, Binyuan Hui, Bo~Zheng, Bowen Yu, Chengyuan Li, Dayiheng Liu, Fei Huang, Haoran Wei, Huan Lin, Jian Yang, Jianhong Tu, Jianwei Zhang, Jianxin Yang, Jiaxi Yang, Jingren Zhou, Junyang Lin, Kai Dang, Keming Lu, Keqin Bao, Kexin Yang, Le~Yu, Mei Li, Mingfeng Xue, Pei Zhang, Qin Zhu, Rui Men, Runji Lin, Tianhao Li, Tianyi Tang, Tingyu Xia, Xingzhang Ren, Xuancheng Ren, Yang Fan, Yang Su, Yichang Zhang, Yu~Wan, Yuqiong Liu, Zeyu Cui, Zhenru Zhang, and Zihan Qiu. 2024.
\newblock \href {https://arxiv.org/abs/2412.15115} {Qwen2.5 technical report}.
\newblock \emph{Preprint}, arXiv:2412.15115.

\bibitem[{Reddy et~al.(2018)Reddy, Rocha, and Esteves}]{reddy2018defactonlp}
Aniketh~Janardhan Reddy, Gil Rocha, and Diego Esteves. 2018.
\newblock Defactonlp: Fact verification using entity recognition, tfidf vector comparison and decomposable attention.
\newblock \emph{arXiv preprint arXiv:1809.00509}.

\bibitem[{Reimers and Gurevych(2019{\natexlab{a}})}]{ref_SBERT}
Nils Reimers and Iryna Gurevych. 2019{\natexlab{a}}.
\newblock \href {https://doi.org/10.18653/v1/D19-1410} {Sentence-{BERT}: Sentence embeddings using {S}iamese {BERT}-networks}.
\newblock In \emph{Proceedings of the 2019 Conference on Empirical Methods in Natural Language Processing and the 9th International Joint Conference on Natural Language Processing (EMNLP-IJCNLP)}, pages 3982--3992, Hong Kong, China. Association for Computational Linguistics.

\bibitem[{Reimers and Gurevych(2019{\natexlab{b}})}]{reimers-gurevych-2019-sentence}
Nils Reimers and Iryna Gurevych. 2019{\natexlab{b}}.
\newblock \href {https://doi.org/10.18653/v1/D19-1410} {Sentence-{BERT}: Sentence embeddings using {S}iamese {BERT}-networks}.
\newblock In \emph{Proceedings of the 2019 Conference on Empirical Methods in Natural Language Processing and the 9th International Joint Conference on Natural Language Processing (EMNLP-IJCNLP)}, pages 3982--3992, Hong Kong, China. Association for Computational Linguistics.

\bibitem[{Schimanski et~al.(2024)Schimanski, Ni, Kraus, Ash, and Leippold}]{schimanski-etal-2024-towards}
Tobias Schimanski, Jingwei Ni, Mathias Kraus, Elliott Ash, and Markus Leippold. 2024.
\newblock \href {https://doi.org/10.18653/v1/2024.acl-long.105} {Towards faithful and robust {LLM} specialists for evidence-based question-answering}.
\newblock In \emph{Proceedings of the 62nd Annual Meeting of the Association for Computational Linguistics (Volume 1: Long Papers)}, pages 1913--1931, Bangkok, Thailand. Association for Computational Linguistics.

\bibitem[{Soleimani et~al.(2020)Soleimani, Monz, and Worring}]{ref_bert_claim_evidence}
Amir Soleimani, Christof Monz, and Marcel Worring. 2020.
\newblock \href {https://doi.org/10.1007/978-3-030-45442-5_45} {Bert for evidence retrieval and claim verification}.
\newblock In \emph{Advances in Information Retrieval: 42nd European Conference on IR Research, ECIR 2020, Lisbon, Portugal, April 14–17, 2020, Proceedings, Part II}, page 359–366, Berlin, Heidelberg. Springer-Verlag.

\bibitem[{Team et~al.(2024)Team, Anil, Borgeaud, Alayrac, Yu, Soricut, Schalkwyk, Dai, Hauth, Millican, Silver, Johnson, Antonoglou, Schrittwieser, Glaese, Chen, Pitler, Lillicrap, Lazaridou, Firat, Molloy, Isard, Barham, Hennigan, Lee, Viola, Reynolds, Xu, Doherty, Collins, Meyer, Rutherford, Moreira, Ayoub, Goel, Krawczyk, Du, Chi, Cheng, Ni, Shah, Kane, Chan, Faruqui, Severyn, Lin, Li, Cheng, Ittycheriah, Mahdieh, Chen, Sun, Tran, Bagri, Lakshminarayanan, Liu, Orban, Güra, Zhou, Song, Boffy, Ganapathy, Zheng, Choe, Ágoston Weisz, Zhu, Lu, Gopal, Kahn, Kula, Pitman, Shah, Taropa, Merey, Baeuml, Chen, Shafey, Zhang, Sercinoglu, Tucker, Piqueras, Krikun, Barr, Savinov, Danihelka, Roelofs, White, Andreassen, von Glehn, Yagati, Kazemi, Gonzalez, Khalman, Sygnowski, Frechette, Smith, Culp, Proleev, Luan, Chen, Lottes, Schucher, Lebron, Rrustemi, Clay, Crone, Kocisky, Zhao, Perz, Yu, Howard, Bloniarz, Rae, Lu, Sifre, Maggioni, Alcober, Garrette, Barnes, Thakoor, Austin, Barth-Maron, Wong, Joshi, Chaabouni,
  Fatiha, Ahuja, Tomar, Senter, Chadwick, Kornakov, Attaluri, Iturrate, Liu, Li, Cogan, Chen, Jia, Gu, Zhang, Grimstad, Hartman, Garcia, Pillai, Devlin, Laskin, de~Las~Casas, Valter, Tao, Blanco, Badia, Reitter, Chen, Brennan, Rivera, Brin, Iqbal, Surita, Labanowski, Rao, Winkler, Parisotto, Gu, Olszewska, Addanki, Miech, Louis, Teplyashin, Brown, Catt, Balaguer, Xiang, Wang, Ashwood, Briukhov, Webson, Ganapathy, Sanghavi, Kannan, Chang, Stjerngren, Djolonga, Sun, Bapna, Aitchison, Pejman, Michalewski, Yu, Wang, Love, Ahn, Bloxwich, Han, Humphreys, Sellam, Bradbury, Godbole, Samangooei, Damoc, Kaskasoli, Arnold, Vasudevan, Agrawal, Riesa, Lepikhin, Tanburn, Srinivasan, Lim, Hodkinson, Shyam, Ferret, Hand, Garg, Paine, Li, Li, Giang, Neitz, Abbas, York, Reid, Cole, Chowdhery, Das, Rogozińska, Nikolaev, Sprechmann, Nado, Zilka, Prost, He, Monteiro, Mishra, Welty, Newlan, Jia, Allamanis, Hu, de~Liedekerke, Gilmer, Saroufim, Rijhwani, Hou, Shrivastava, Baddepudi, Goldin, Ozturel, Cassirer, Xu, Sohn, Sachan,
  Amplayo, Swanson, Petrova, Narayan, Guez, Brahma, Landon, Patel, Zhao, Villela, Wang, Jia, Rahtz, Giménez, Yeung, Keeling, Georgiev, Mincu, Wu, Haykal, Saputro, Vodrahalli, Qin, Cankara, Sharma, Fernando, Hawkins, Neyshabur, Kim, Hutter, Agrawal, Castro-Ros, van~den Driessche, Wang, Yang, yiin Chang, Komarek, McIlroy, Lučić, Zhang, Farhan, Sharman, Natsev, Michel, Bansal, Qiao, Cao, Shakeri, Butterfield, Chung, Rubenstein, Agrawal, Mensch, Soparkar, Lenc, Chung, Pope, Maggiore, Kay, Jhakra, Wang, Maynez, Phuong, Tobin, Tacchetti, Trebacz, Robinson, Katariya, Riedel, Bailey, Xiao, Ghelani, Aroyo, Slone, Houlsby, Xiong, Yang, Gribovskaya, Adler, Wirth, Lee, Li, Kagohara, Pavagadhi, Bridgers, Bortsova, Ghemawat, Ahmed, Liu, Powell, Bolina, Iinuma, Zablotskaia, Besley, Chung, Dozat, Comanescu, Si, Greer, Su, Polacek, Kaufman, Tokumine, Hu, Buchatskaya, Miao, Elhawaty, Siddhant, Tomasev, Xing, Greer, Miller, Ashraf, Roy, Zhang, Ma, Filos, Besta, Blevins, Klimenko, Yeh, Changpinyo, Mu, Chang, Pajarskas, Muir,
  Cohen, Lan, Haridasan, Marathe, Hansen, Douglas, Samuel, Wang, Austin, Lan, Jiang, Chiu, Lorenzo, Sjösund, Cevey, Gleicher, Avrahami, Boral, Srinivasan, Selo, May, Aisopos, Hussenot, Soares, Baumli, Chang, Recasens, Caine, Pritzel, Pavetic, Pardo, Gergely, Frye, Ramasesh, Horgan, Badola, Kassner, Roy, Dyer, Campos, Tomala, Tang, Badawy, White, Mustafa, Lang, Jindal, Vikram, Gong, Caelles, Hemsley, Thornton, Feng, Stokowiec, Zheng, Thacker, Çağlar Ünlü, Zhang, Saleh, Svensson, Bileschi, Patil, Anand, Ring, Tsihlas, Vezer, Selvi, Shevlane, Rodriguez, Kwiatkowski, Daruki, Rong, Dafoe, FitzGerald, Gu-Lemberg, Khan, Hendricks, Pellat, Feinberg, Cobon-Kerr, Sainath, Rauh, Hashemi, Ives, Hasson, Noland, Cao, Byrd, Hou, Wang, Sottiaux, Paganini, Lespiau, Moufarek, Hassan, Shivakumar, van Amersfoort, Mandhane, Joshi, Goyal, Tung, Brock, Sheahan, Misra, Li, Rakićević, Dehghani, Liu, Mittal, Oh, Noury, Sezener, Huot, Lamm, Cao, Chen, Mudgal, Stella, Brooks, Vasudevan, Liu, Chain, Melinkeri, Cohen, Wang,
  Seymore, Zubkov, Goel, Yue, Krishnakumaran, Albert, Hurley, Sano, Mohananey, Joughin, Filonov, Tomasz, Eldawy, Lim, Rishi, Badiezadegan, Bos, Chang, Jain, Padmanabhan, Puttagunta, Krishna, Baker, Kalb, Bedapudi, Kurzrok, Lei, Yu, Litvin, Zhou, Wu, Sobell, Siciliano, Papir, Neale, Bragagnolo, Toor, Chen, Anklin, Wang, Feng, Gholami, Ling, Liu, Walter, Moghaddam, Kishore, Adamek, Mercado, Mallinson, Wandekar, Cagle, Ofek, Garrido, Lombriser, Mukha, Sun, Mohammad, Matak, Qian, Peswani, Janus, Yuan, Schelin, David, Garg, He, Duzhyi, Älgmyr, Lottaz, Li, Yadav, Xu, Chinien, Shivanna, Chuklin, Li, Spadine, Wolfe, Mohamed, Das, Dai, He, von Dincklage, Upadhyay, Maurya, Chi, Krause, Salama, Rabinovitch, M, Selvan, Dektiarev, Ghiasi, Guven, Gupta, Liu, Sharma, Shtacher, Paul, Akerlund, Aubet, Huang, Zhu, Zhu, Teixeira, Fritze, Bertolini, Marinescu, Bölle, Paulus, Gupta, Latkar, Chang, Sanders, Wilson, Wu, Tan, Thiet, Doshi, Lall, Mishra, Chen, Luong, Benjamin, Lee, Andrejczuk, Rabiej, Ranjan, Styrc, Yin, Simon,
  Harriott, Bansal, Robsky, Bacon, Greene, Mirylenka, Zhou, Sarvana, Goyal, Andermatt, Siegler, Horn, Israel, Pongetti, Chen, Selvatici, Silva, Wang, Tolins, Guu, Yogev, Cai, Agostini, Shah, Nguyen, Donnaile, Pereira, Friso, Stambler, Kurzrok, Kuang, Romanikhin, Geller, Yan, Jang, Lee, Fica, Malmi, Tan, Banica, Balle, Pham, Huang, Avram, Shi, Singh, Hidey, Ahuja, Saxena, Dooley, Potharaju, O'Neill, Gokulchandran, Foley, Zhao, Dusenberry, Liu, Mehta, Kotikalapudi, Safranek-Shrader, Goodman, Kessinger, Globen, Kolhar, Gorgolewski, Ibrahim, Song, Eichenbaum, Brovelli, Potluri, Lahoti, Baetu, Ghorbani, Chen, Crawford, Pal, Sridhar, Gurita, Mujika, Petrovski, Cedoz, Li, Chen, Santo, Goyal, Punjabi, Kappaganthu, Kwak, LV, Velury, Choudhury, Hall, Shah, Figueira, Thomas, Lu, Zhou, Kumar, Jurdi, Chikkerur, Ma, Yu, Kwak, Ähdel, Rajayogam, Choma, Liu, Barua, Ji, Park, Hellendoorn, Bailey, Bilal, Zhou, Khatir, Sutton, Rzadkowski, Macintosh, Shagin, Medina, Liang, Zhou, Shah, Bi, Dankovics, Banga, Lehmann, Bredesen,
  Lin, Hoffmann, Lai, Chung, Yang, Balani, Bražinskas, Sozanschi, Hayes, Alcalde, Makarov, Chen, Stella, Snijders, Mandl, Kärrman, Nowak, Wu, Dyck, Vaidyanathan, R, Mallet, Rudominer, Johnston, Mittal, Udathu, Christensen, Verma, Irving, Santucci, Elsayed, Davoodi, Georgiev, Tenney, Hua, Cideron, Leurent, Alnahlawi, Georgescu, Wei, Zheng, Scandinaro, Jiang, Snoek, Sundararajan, Wang, Ontiveros, Karo, Cole, Rajashekhar, Tumeh, Ben-David, Jain, Uesato, Datta, Bunyan, Wu, Zhang, Stanczyk, Zhang, Steiner, Naskar, Azzam, Johnson, Paszke, Chiu, Elias, Mohiuddin, Muhammad, Miao, Lee, Vieillard, Park, Zhang, Stanway, Garmon, Karmarkar, Dong, Lee, Kumar, Zhou, Evens, Isaac, Irving, Loper, Fink, Arkatkar, Chen, Shafran, Petrychenko, Chen, Jia, Levskaya, Zhu, Grabowski, Mao, Magni, Yao, Snaider, Casagrande, Palmer, Suganthan, Castaño, Giannoumis, Kim, Rybiński, Sreevatsa, Prendki, Soergel, Goedeckemeyer, Gierke, Jafari, Gaba, Wiesner, Wright, Wei, Vashisht, Kulizhskaya, Hoover, Le, Li, Iwuanyanwu, Liu, Ramirez,
  Khorlin, Cui, LIN, Wu, Aguilar, Pallo, Chakladar, Perng, Abellan, Zhang, Dasgupta, Kushman, Penchev, Repina, Wu, van~der Weide, Ponnapalli, Kaplan, Simsa, Li, Dousse, Yang, Piper, Ie, Pasumarthi, Lintz, Vijayakumar, Andor, Valenzuela, Lui, Paduraru, Peng, Lee, Zhang, Greene, Nguyen, Kurylowicz, Hardin, Dixon, Janzer, Choo, Feng, Zhang, Singhal, Du, McKinnon, Antropova, Bolukbasi, Keller, Reid, Finchelstein, Raad, Crocker, Hawkins, Dadashi, Gaffney, Franko, Bulanova, Leblond, Chung, Askham, Cobo, Xu, Fischer, Xu, Sorokin, Alberti, Lin, Evans, Dimitriev, Forbes, Banarse, Tung, Omernick, Bishop, Sterneck, Jain, Xia, Amid, Piccinno, Wang, Banzal, Mankowitz, Polozov, Krakovna, Brown, Bateni, Duan, Firoiu, Thotakuri, Natan, Geist, tan Girgin, Li, Ye, Roval, Tojo, Kwong, Lee-Thorp, Yew, Sinopalnikov, Ramos, Mellor, Sharma, Wu, Miller, Sonnerat, Vnukov, Greig, Beattie, Caveness, Bai, Eisenschlos, Korchemniy, Tsai, Jasarevic, Kong, Dao, Zheng, Liu, Yang, Zhu, Teh, Sanmiya, Gladchenko, Trdin, Toyama, Rosen, Tavakkol,
  Xue, Elkind, Woodman, Carpenter, Papamakarios, Kemp, Kafle, Grunina, Sinha, Talbert, Wu, Owusu-Afriyie, Du, Thornton, Pont-Tuset, Narayana, Li, Fatehi, Wieting, Ajmeri, Uria, Ko, Knight, Héliou, Niu, Gu, Pang, Li, Levine, Stolovich, Santamaria-Fernandez, Goenka, Yustalim, Strudel, Elqursh, Deck, Lee, Li, Levin, Hoffmann, Holtmann-Rice, Bachem, Arora, Koh, Yeganeh, Põder, Tariq, Sun, Ionita, Seyedhosseini, Tafti, Liu, Gulati, Liu, Ye, Chrzaszcz, Wang, Sethi, Li, Brown, Singh, Fan, Parisi, Stanton, Koverkathu, Choquette-Choo, Li, Lu, Ittycheriah, Shroff, Varadarajan, Bahargam, Willoughby, Gaddy, Desjardins, Cornero, Robenek, Mittal, Albrecht, Shenoy, Moiseev, Jacobsson, Ghaffarkhah, Rivière, Walton, Crepy, Parrish, Zhou, Farabet, Radebaugh, Srinivasan, van~der Salm, Fidjeland, Scellato, Latorre-Chimoto, Klimczak-Plucińska, Bridson, de~Cesare, Hudson, Mendolicchio, Walker, Morris, Mauger, Guseynov, Reid, Odoom, Loher, Cotruta, Yenugula, Grewe, Petrushkina, Duerig, Sanchez, Yadlowsky, Shen, Globerson, Webb,
  Dua, Li, Bhupatiraju, Hurt, Qureshi, Agarwal, Shani, Eyal, Khare, Belle, Wang, Tekur, Kale, Wei, Sang, Saeta, Liechty, Sun, Zhao, Lee, Nayak, Fritz, Vuyyuru, Aslanides, Vyas, Wicke, Ma, Eltyshev, Martin, Cate, Manyika, Amiri, Kim, Xiong, Kang, Luisier, Tripuraneni, Madras, Guo, Waters, Wang, Ainslie, Baldridge, Zhang, Pruthi, Bauer, Yang, Mansour, Gelman, Xu, Polovets, Liu, Cai, Chen, Sheng, Xue, Ozair, Angermueller, Li, Sinha, Wang, Wiesinger, Koukoumidis, Tian, Iyer, Gurumurthy, Goldenson, Shah, Blake, Yu, Urbanowicz, Palomaki, Fernando, Durden, Mehta, Momchev, Rahimtoroghi, Georgaki, Raul, Ruder, Redshaw, Lee, Zhou, Jalan, Li, Hechtman, Schuh, Nasr, Milan, Mikulik, Franco, Green, Nguyen, Kelley, Mahendru, Hu, Howland, Vargas, Hui, Bansal, Rao, Ghiya, Wang, Ye, Sarr, Preston, Elish, Li, Kaku, Gupta, Pasupat, Juan, Someswar, M., Chen, Amini, Fabrikant, Chu, Dong, Muthal, Buthpitiya, Jauhari, Hua, Khandelwal, Hitron, Ren, Rinaldi, Drath, Dabush, Jiang, Godhia, Sachs, Chen, Fan, Taitelbaum, Noga, Dai, Wang,
  Liang, Hamer, Ferng, Elkind, Atias, Lee, Listík, Carlen, van~de Kerkhof, Pikus, Zaher, Müller, Zykova, Stefanec, Gatsko, Hirnschall, Sethi, Xu, Ahuja, Tsai, Stefanoiu, Feng, Dhandhania, Katyal, Gupta, Parulekar, Pitta, Zhao, Bhatia, Bhavnani, Alhadlaq, Li, Danenberg, Tu, Pine, Filippova, Ghosh, Limonchik, Urala, Lanka, Clive, Sun, Li, Wu, Hongtongsak, Li, Thakkar, Omarov, Majmundar, Alverson, Kucharski, Patel, Jain, Zabelin, Pelagatti, Kohli, Kumar, Kim, Sankar, Shah, Ramachandruni, Zeng, Bariach, Weidinger, Vu, Andreev, He, Hui, Kashem, Subramanya, Hsiao, Hassabis, Kavukcuoglu, Sadovsky, Le, Strohman, Wu, Petrov, Dean, and Vinyals}]{geminiteam2024geminifamilyhighlycapable}
Gemini Team, Rohan Anil, Sebastian Borgeaud, Jean-Baptiste Alayrac, Jiahui Yu, Radu Soricut, Johan Schalkwyk, Andrew~M. Dai, Anja Hauth, Katie Millican, David Silver, Melvin Johnson, Ioannis Antonoglou, Julian Schrittwieser, Amelia Glaese, Jilin Chen, Emily Pitler, Timothy Lillicrap, Angeliki Lazaridou, Orhan Firat, James Molloy, Michael Isard, Paul~R. Barham, Tom Hennigan, Benjamin Lee, Fabio Viola, Malcolm Reynolds, Yuanzhong Xu, Ryan Doherty, Eli Collins, Clemens Meyer, Eliza Rutherford, Erica Moreira, Kareem Ayoub, Megha Goel, Jack Krawczyk, Cosmo Du, Ed~Chi, Heng-Tze Cheng, Eric Ni, Purvi Shah, Patrick Kane, Betty Chan, Manaal Faruqui, Aliaksei Severyn, Hanzhao Lin, YaGuang Li, Yong Cheng, Abe Ittycheriah, Mahdis Mahdieh, Mia Chen, Pei Sun, Dustin Tran, Sumit Bagri, Balaji Lakshminarayanan, Jeremiah Liu, Andras Orban, Fabian Güra, Hao Zhou, Xinying Song, Aurelien Boffy, Harish Ganapathy, Steven Zheng, HyunJeong Choe, Ágoston Weisz, Tao Zhu, Yifeng Lu, Siddharth Gopal, Jarrod Kahn, Maciej Kula, Jeff
  Pitman, Rushin Shah, Emanuel Taropa, Majd~Al Merey, Martin Baeuml, Zhifeng Chen, Laurent~El Shafey, Yujing Zhang, Olcan Sercinoglu, George Tucker, Enrique Piqueras, Maxim Krikun, Iain Barr, Nikolay Savinov, Ivo Danihelka, Becca Roelofs, Anaïs White, Anders Andreassen, Tamara von Glehn, Lakshman Yagati, Mehran Kazemi, Lucas Gonzalez, Misha Khalman, Jakub Sygnowski, Alexandre Frechette, Charlotte Smith, Laura Culp, Lev Proleev, Yi~Luan, Xi~Chen, James Lottes, Nathan Schucher, Federico Lebron, Alban Rrustemi, Natalie Clay, Phil Crone, Tomas Kocisky, Jeffrey Zhao, Bartek Perz, Dian Yu, Heidi Howard, Adam Bloniarz, Jack~W. Rae, Han Lu, Laurent Sifre, Marcello Maggioni, Fred Alcober, Dan Garrette, Megan Barnes, Shantanu Thakoor, Jacob Austin, Gabriel Barth-Maron, William Wong, Rishabh Joshi, Rahma Chaabouni, Deeni Fatiha, Arun Ahuja, Gaurav~Singh Tomar, Evan Senter, Martin Chadwick, Ilya Kornakov, Nithya Attaluri, Iñaki Iturrate, Ruibo Liu, Yunxuan Li, Sarah Cogan, Jeremy Chen, Chao Jia, Chenjie Gu, Qiao Zhang,
  Jordan Grimstad, Ale~Jakse Hartman, Xavier Garcia, Thanumalayan~Sankaranarayana Pillai, Jacob Devlin, Michael Laskin, Diego de~Las~Casas, Dasha Valter, Connie Tao, Lorenzo Blanco, Adrià~Puigdomènech Badia, David Reitter, Mianna Chen, Jenny Brennan, Clara Rivera, Sergey Brin, Shariq Iqbal, Gabriela Surita, Jane Labanowski, Abhi Rao, Stephanie Winkler, Emilio Parisotto, Yiming Gu, Kate Olszewska, Ravi Addanki, Antoine Miech, Annie Louis, Denis Teplyashin, Geoff Brown, Elliot Catt, Jan Balaguer, Jackie Xiang, Pidong Wang, Zoe Ashwood, Anton Briukhov, Albert Webson, Sanjay Ganapathy, Smit Sanghavi, Ajay Kannan, Ming-Wei Chang, Axel Stjerngren, Josip Djolonga, Yuting Sun, Ankur Bapna, Matthew Aitchison, Pedram Pejman, Henryk Michalewski, Tianhe Yu, Cindy Wang, Juliette Love, Junwhan Ahn, Dawn Bloxwich, Kehang Han, Peter Humphreys, Thibault Sellam, James Bradbury, Varun Godbole, Sina Samangooei, Bogdan Damoc, Alex Kaskasoli, Sébastien M.~R. Arnold, Vijay Vasudevan, Shubham Agrawal, Jason Riesa, Dmitry
  Lepikhin, Richard Tanburn, Srivatsan Srinivasan, Hyeontaek Lim, Sarah Hodkinson, Pranav Shyam, Johan Ferret, Steven Hand, Ankush Garg, Tom~Le Paine, Jian Li, Yujia Li, Minh Giang, Alexander Neitz, Zaheer Abbas, Sarah York, Machel Reid, Elizabeth Cole, Aakanksha Chowdhery, Dipanjan Das, Dominika Rogozińska, Vitaliy Nikolaev, Pablo Sprechmann, Zachary Nado, Lukas Zilka, Flavien Prost, Luheng He, Marianne Monteiro, Gaurav Mishra, Chris Welty, Josh Newlan, Dawei Jia, Miltiadis Allamanis, Clara~Huiyi Hu, Raoul de~Liedekerke, Justin Gilmer, Carl Saroufim, Shruti Rijhwani, Shaobo Hou, Disha Shrivastava, Anirudh Baddepudi, Alex Goldin, Adnan Ozturel, Albin Cassirer, Yunhan Xu, Daniel Sohn, Devendra Sachan, Reinald~Kim Amplayo, Craig Swanson, Dessie Petrova, Shashi Narayan, Arthur Guez, Siddhartha Brahma, Jessica Landon, Miteyan Patel, Ruizhe Zhao, Kevin Villela, Luyu Wang, Wenhao Jia, Matthew Rahtz, Mai Giménez, Legg Yeung, James Keeling, Petko Georgiev, Diana Mincu, Boxi Wu, Salem Haykal, Rachel Saputro, Kiran
  Vodrahalli, James Qin, Zeynep Cankara, Abhanshu Sharma, Nick Fernando, Will Hawkins, Behnam Neyshabur, Solomon Kim, Adrian Hutter, Priyanka Agrawal, Alex Castro-Ros, George van~den Driessche, Tao Wang, Fan Yang, Shuo yiin Chang, Paul Komarek, Ross McIlroy, Mario Lučić, Guodong Zhang, Wael Farhan, Michael Sharman, Paul Natsev, Paul Michel, Yamini Bansal, Siyuan Qiao, Kris Cao, Siamak Shakeri, Christina Butterfield, Justin Chung, Paul~Kishan Rubenstein, Shivani Agrawal, Arthur Mensch, Kedar Soparkar, Karel Lenc, Timothy Chung, Aedan Pope, Loren Maggiore, Jackie Kay, Priya Jhakra, Shibo Wang, Joshua Maynez, Mary Phuong, Taylor Tobin, Andrea Tacchetti, Maja Trebacz, Kevin Robinson, Yash Katariya, Sebastian Riedel, Paige Bailey, Kefan Xiao, Nimesh Ghelani, Lora Aroyo, Ambrose Slone, Neil Houlsby, Xuehan Xiong, Zhen Yang, Elena Gribovskaya, Jonas Adler, Mateo Wirth, Lisa Lee, Music Li, Thais Kagohara, Jay Pavagadhi, Sophie Bridgers, Anna Bortsova, Sanjay Ghemawat, Zafarali Ahmed, Tianqi Liu, Richard Powell,
  Vijay Bolina, Mariko Iinuma, Polina Zablotskaia, James Besley, Da-Woon Chung, Timothy Dozat, Ramona Comanescu, Xiance Si, Jeremy Greer, Guolong Su, Martin Polacek, Raphaël~Lopez Kaufman, Simon Tokumine, Hexiang Hu, Elena Buchatskaya, Yingjie Miao, Mohamed Elhawaty, Aditya Siddhant, Nenad Tomasev, Jinwei Xing, Christina Greer, Helen Miller, Shereen Ashraf, Aurko Roy, Zizhao Zhang, Ada Ma, Angelos Filos, Milos Besta, Rory Blevins, Ted Klimenko, Chih-Kuan Yeh, Soravit Changpinyo, Jiaqi Mu, Oscar Chang, Mantas Pajarskas, Carrie Muir, Vered Cohen, Charline~Le Lan, Krishna Haridasan, Amit Marathe, Steven Hansen, Sholto Douglas, Rajkumar Samuel, Mingqiu Wang, Sophia Austin, Chang Lan, Jiepu Jiang, Justin Chiu, Jaime~Alonso Lorenzo, Lars~Lowe Sjösund, Sébastien Cevey, Zach Gleicher, Thi Avrahami, Anudhyan Boral, Hansa Srinivasan, Vittorio Selo, Rhys May, Konstantinos Aisopos, Léonard Hussenot, Livio~Baldini Soares, Kate Baumli, Michael~B. Chang, Adrià Recasens, Ben Caine, Alexander Pritzel, Filip Pavetic,
  Fabio Pardo, Anita Gergely, Justin Frye, Vinay Ramasesh, Dan Horgan, Kartikeya Badola, Nora Kassner, Subhrajit Roy, Ethan Dyer, Víctor~Campos Campos, Alex Tomala, Yunhao Tang, Dalia~El Badawy, Elspeth White, Basil Mustafa, Oran Lang, Abhishek Jindal, Sharad Vikram, Zhitao Gong, Sergi Caelles, Ross Hemsley, Gregory Thornton, Fangxiaoyu Feng, Wojciech Stokowiec, Ce~Zheng, Phoebe Thacker, Çağlar Ünlü, Zhishuai Zhang, Mohammad Saleh, James Svensson, Max Bileschi, Piyush Patil, Ankesh Anand, Roman Ring, Katerina Tsihlas, Arpi Vezer, Marco Selvi, Toby Shevlane, Mikel Rodriguez, Tom Kwiatkowski, Samira Daruki, Keran Rong, Allan Dafoe, Nicholas FitzGerald, Keren Gu-Lemberg, Mina Khan, Lisa~Anne Hendricks, Marie Pellat, Vladimir Feinberg, James Cobon-Kerr, Tara Sainath, Maribeth Rauh, Sayed~Hadi Hashemi, Richard Ives, Yana Hasson, Eric Noland, Yuan Cao, Nathan Byrd, Le~Hou, Qingze Wang, Thibault Sottiaux, Michela Paganini, Jean-Baptiste Lespiau, Alexandre Moufarek, Samer Hassan, Kaushik Shivakumar, Joost van
  Amersfoort, Amol Mandhane, Pratik Joshi, Anirudh Goyal, Matthew Tung, Andrew Brock, Hannah Sheahan, Vedant Misra, Cheng Li, Nemanja Rakićević, Mostafa Dehghani, Fangyu Liu, Sid Mittal, Junhyuk Oh, Seb Noury, Eren Sezener, Fantine Huot, Matthew Lamm, Nicola~De Cao, Charlie Chen, Sidharth Mudgal, Romina Stella, Kevin Brooks, Gautam Vasudevan, Chenxi Liu, Mainak Chain, Nivedita Melinkeri, Aaron Cohen, Venus Wang, Kristie Seymore, Sergey Zubkov, Rahul Goel, Summer Yue, Sai Krishnakumaran, Brian Albert, Nate Hurley, Motoki Sano, Anhad Mohananey, Jonah Joughin, Egor Filonov, Tomasz, Yomna Eldawy, Jiawern Lim, Rahul Rishi, Shirin Badiezadegan, Taylor Bos, Jerry Chang, Sanil Jain, Sri Gayatri~Sundara Padmanabhan, Subha Puttagunta, Kalpesh Krishna, Leslie Baker, Norbert Kalb, Vamsi Bedapudi, Adam Kurzrok, Shuntong Lei, Anthony Yu, Oren Litvin, Xiang Zhou, Zhichun Wu, Sam Sobell, Andrea Siciliano, Alan Papir, Robby Neale, Jonas Bragagnolo, Tej Toor, Tina Chen, Valentin Anklin, Feiran Wang, Richie Feng, Milad
  Gholami, Kevin Ling, Lijuan Liu, Jules Walter, Hamid Moghaddam, Arun Kishore, Jakub Adamek, Tyler Mercado, Jonathan Mallinson, Siddhinita Wandekar, Stephen Cagle, Eran Ofek, Guillermo Garrido, Clemens Lombriser, Maksim Mukha, Botu Sun, Hafeezul~Rahman Mohammad, Josip Matak, Yadi Qian, Vikas Peswani, Pawel Janus, Quan Yuan, Leif Schelin, Oana David, Ankur Garg, Yifan He, Oleksii Duzhyi, Anton Älgmyr, Timothée Lottaz, Qi~Li, Vikas Yadav, Luyao Xu, Alex Chinien, Rakesh Shivanna, Aleksandr Chuklin, Josie Li, Carrie Spadine, Travis Wolfe, Kareem Mohamed, Subhabrata Das, Zihang Dai, Kyle He, Daniel von Dincklage, Shyam Upadhyay, Akanksha Maurya, Luyan Chi, Sebastian Krause, Khalid Salama, Pam~G Rabinovitch, Pavan Kumar~Reddy M, Aarush Selvan, Mikhail Dektiarev, Golnaz Ghiasi, Erdem Guven, Himanshu Gupta, Boyi Liu, Deepak Sharma, Idan~Heimlich Shtacher, Shachi Paul, Oscar Akerlund, François-Xavier Aubet, Terry Huang, Chen Zhu, Eric Zhu, Elico Teixeira, Matthew Fritze, Francesco Bertolini, Liana-Eleonora
  Marinescu, Martin Bölle, Dominik Paulus, Khyatti Gupta, Tejasi Latkar, Max Chang, Jason Sanders, Roopa Wilson, Xuewei Wu, Yi-Xuan Tan, Lam~Nguyen Thiet, Tulsee Doshi, Sid Lall, Swaroop Mishra, Wanming Chen, Thang Luong, Seth Benjamin, Jasmine Lee, Ewa Andrejczuk, Dominik Rabiej, Vipul Ranjan, Krzysztof Styrc, Pengcheng Yin, Jon Simon, Malcolm~Rose Harriott, Mudit Bansal, Alexei Robsky, Geoff Bacon, David Greene, Daniil Mirylenka, Chen Zhou, Obaid Sarvana, Abhimanyu Goyal, Samuel Andermatt, Patrick Siegler, Ben Horn, Assaf Israel, Francesco Pongetti, Chih-Wei~"Louis" Chen, Marco Selvatici, Pedro Silva, Kathie Wang, Jackson Tolins, Kelvin Guu, Roey Yogev, Xiaochen Cai, Alessandro Agostini, Maulik Shah, Hung Nguyen, Noah~Ó Donnaile, Sébastien Pereira, Linda Friso, Adam Stambler, Adam Kurzrok, Chenkai Kuang, Yan Romanikhin, Mark Geller, ZJ~Yan, Kane Jang, Cheng-Chun Lee, Wojciech Fica, Eric Malmi, Qijun Tan, Dan Banica, Daniel Balle, Ryan Pham, Yanping Huang, Diana Avram, Hongzhi Shi, Jasjot Singh, Chris
  Hidey, Niharika Ahuja, Pranab Saxena, Dan Dooley, Srividya~Pranavi Potharaju, Eileen O'Neill, Anand Gokulchandran, Ryan Foley, Kai Zhao, Mike Dusenberry, Yuan Liu, Pulkit Mehta, Ragha Kotikalapudi, Chalence Safranek-Shrader, Andrew Goodman, Joshua Kessinger, Eran Globen, Prateek Kolhar, Chris Gorgolewski, Ali Ibrahim, Yang Song, Ali Eichenbaum, Thomas Brovelli, Sahitya Potluri, Preethi Lahoti, Cip Baetu, Ali Ghorbani, Charles Chen, Andy Crawford, Shalini Pal, Mukund Sridhar, Petru Gurita, Asier Mujika, Igor Petrovski, Pierre-Louis Cedoz, Chenmei Li, Shiyuan Chen, Niccolò~Dal Santo, Siddharth Goyal, Jitesh Punjabi, Karthik Kappaganthu, Chester Kwak, Pallavi LV, Sarmishta Velury, Himadri Choudhury, Jamie Hall, Premal Shah, Ricardo Figueira, Matt Thomas, Minjie Lu, Ting Zhou, Chintu Kumar, Thomas Jurdi, Sharat Chikkerur, Yenai Ma, Adams Yu, Soo Kwak, Victor Ähdel, Sujeevan Rajayogam, Travis Choma, Fei Liu, Aditya Barua, Colin Ji, Ji~Ho Park, Vincent Hellendoorn, Alex Bailey, Taylan Bilal, Huanjie Zhou,
  Mehrdad Khatir, Charles Sutton, Wojciech Rzadkowski, Fiona Macintosh, Konstantin Shagin, Paul Medina, Chen Liang, Jinjing Zhou, Pararth Shah, Yingying Bi, Attila Dankovics, Shipra Banga, Sabine Lehmann, Marissa Bredesen, Zifan Lin, John~Eric Hoffmann, Jonathan Lai, Raynald Chung, Kai Yang, Nihal Balani, Arthur Bražinskas, Andrei Sozanschi, Matthew Hayes, Héctor~Fernández Alcalde, Peter Makarov, Will Chen, Antonio Stella, Liselotte Snijders, Michael Mandl, Ante Kärrman, Paweł Nowak, Xinyi Wu, Alex Dyck, Krishnan Vaidyanathan, Raghavender R, Jessica Mallet, Mitch Rudominer, Eric Johnston, Sushil Mittal, Akhil Udathu, Janara Christensen, Vishal Verma, Zach Irving, Andreas Santucci, Gamaleldin Elsayed, Elnaz Davoodi, Marin Georgiev, Ian Tenney, Nan Hua, Geoffrey Cideron, Edouard Leurent, Mahmoud Alnahlawi, Ionut Georgescu, Nan Wei, Ivy Zheng, Dylan Scandinaro, Heinrich Jiang, Jasper Snoek, Mukund Sundararajan, Xuezhi Wang, Zack Ontiveros, Itay Karo, Jeremy Cole, Vinu Rajashekhar, Lara Tumeh, Eyal
  Ben-David, Rishub Jain, Jonathan Uesato, Romina Datta, Oskar Bunyan, Shimu Wu, John Zhang, Piotr Stanczyk, Ye~Zhang, David Steiner, Subhajit Naskar, Michael Azzam, Matthew Johnson, Adam Paszke, Chung-Cheng Chiu, Jaume~Sanchez Elias, Afroz Mohiuddin, Faizan Muhammad, Jin Miao, Andrew Lee, Nino Vieillard, Jane Park, Jiageng Zhang, Jeff Stanway, Drew Garmon, Abhijit Karmarkar, Zhe Dong, Jong Lee, Aviral Kumar, Luowei Zhou, Jonathan Evens, William Isaac, Geoffrey Irving, Edward Loper, Michael Fink, Isha Arkatkar, Nanxin Chen, Izhak Shafran, Ivan Petrychenko, Zhe Chen, Johnson Jia, Anselm Levskaya, Zhenkai Zhu, Peter Grabowski, Yu~Mao, Alberto Magni, Kaisheng Yao, Javier Snaider, Norman Casagrande, Evan Palmer, Paul Suganthan, Alfonso Castaño, Irene Giannoumis, Wooyeol Kim, Mikołaj Rybiński, Ashwin Sreevatsa, Jennifer Prendki, David Soergel, Adrian Goedeckemeyer, Willi Gierke, Mohsen Jafari, Meenu Gaba, Jeremy Wiesner, Diana~Gage Wright, Yawen Wei, Harsha Vashisht, Yana Kulizhskaya, Jay Hoover, Maigo Le,
  Lu~Li, Chimezie Iwuanyanwu, Lu~Liu, Kevin Ramirez, Andrey Khorlin, Albert Cui, Tian LIN, Marcus Wu, Ricardo Aguilar, Keith Pallo, Abhishek Chakladar, Ginger Perng, Elena~Allica Abellan, Mingyang Zhang, Ishita Dasgupta, Nate Kushman, Ivo Penchev, Alena Repina, Xihui Wu, Tom van~der Weide, Priya Ponnapalli, Caroline Kaplan, Jiri Simsa, Shuangfeng Li, Olivier Dousse, Fan Yang, Jeff Piper, Nathan Ie, Rama Pasumarthi, Nathan Lintz, Anitha Vijayakumar, Daniel Andor, Pedro Valenzuela, Minnie Lui, Cosmin Paduraru, Daiyi Peng, Katherine Lee, Shuyuan Zhang, Somer Greene, Duc~Dung Nguyen, Paula Kurylowicz, Cassidy Hardin, Lucas Dixon, Lili Janzer, Kiam Choo, Ziqiang Feng, Biao Zhang, Achintya Singhal, Dayou Du, Dan McKinnon, Natasha Antropova, Tolga Bolukbasi, Orgad Keller, David Reid, Daniel Finchelstein, Maria~Abi Raad, Remi Crocker, Peter Hawkins, Robert Dadashi, Colin Gaffney, Ken Franko, Anna Bulanova, Rémi Leblond, Shirley Chung, Harry Askham, Luis~C. Cobo, Kelvin Xu, Felix Fischer, Jun Xu, Christina Sorokin,
  Chris Alberti, Chu-Cheng Lin, Colin Evans, Alek Dimitriev, Hannah Forbes, Dylan Banarse, Zora Tung, Mark Omernick, Colton Bishop, Rachel Sterneck, Rohan Jain, Jiawei Xia, Ehsan Amid, Francesco Piccinno, Xingyu Wang, Praseem Banzal, Daniel~J. Mankowitz, Alex Polozov, Victoria Krakovna, Sasha Brown, MohammadHossein Bateni, Dennis Duan, Vlad Firoiu, Meghana Thotakuri, Tom Natan, Matthieu Geist, Ser tan Girgin, Hui Li, Jiayu Ye, Ofir Roval, Reiko Tojo, Michael Kwong, James Lee-Thorp, Christopher Yew, Danila Sinopalnikov, Sabela Ramos, John Mellor, Abhishek Sharma, Kathy Wu, David Miller, Nicolas Sonnerat, Denis Vnukov, Rory Greig, Jennifer Beattie, Emily Caveness, Libin Bai, Julian Eisenschlos, Alex Korchemniy, Tomy Tsai, Mimi Jasarevic, Weize Kong, Phuong Dao, Zeyu Zheng, Frederick Liu, Fan Yang, Rui Zhu, Tian~Huey Teh, Jason Sanmiya, Evgeny Gladchenko, Nejc Trdin, Daniel Toyama, Evan Rosen, Sasan Tavakkol, Linting Xue, Chen Elkind, Oliver Woodman, John Carpenter, George Papamakarios, Rupert Kemp, Sushant
  Kafle, Tanya Grunina, Rishika Sinha, Alice Talbert, Diane Wu, Denese Owusu-Afriyie, Cosmo Du, Chloe Thornton, Jordi Pont-Tuset, Pradyumna Narayana, Jing Li, Saaber Fatehi, John Wieting, Omar Ajmeri, Benigno Uria, Yeongil Ko, Laura Knight, Amélie Héliou, Ning Niu, Shane Gu, Chenxi Pang, Yeqing Li, Nir Levine, Ariel Stolovich, Rebeca Santamaria-Fernandez, Sonam Goenka, Wenny Yustalim, Robin Strudel, Ali Elqursh, Charlie Deck, Hyo Lee, Zonglin Li, Kyle Levin, Raphael Hoffmann, Dan Holtmann-Rice, Olivier Bachem, Sho Arora, Christy Koh, Soheil~Hassas Yeganeh, Siim Põder, Mukarram Tariq, Yanhua Sun, Lucian Ionita, Mojtaba Seyedhosseini, Pouya Tafti, Zhiyu Liu, Anmol Gulati, Jasmine Liu, Xinyu Ye, Bart Chrzaszcz, Lily Wang, Nikhil Sethi, Tianrun Li, Ben Brown, Shreya Singh, Wei Fan, Aaron Parisi, Joe Stanton, Vinod Koverkathu, Christopher~A. Choquette-Choo, Yunjie Li, TJ~Lu, Abe Ittycheriah, Prakash Shroff, Mani Varadarajan, Sanaz Bahargam, Rob Willoughby, David Gaddy, Guillaume Desjardins, Marco Cornero, Brona
  Robenek, Bhavishya Mittal, Ben Albrecht, Ashish Shenoy, Fedor Moiseev, Henrik Jacobsson, Alireza Ghaffarkhah, Morgane Rivière, Alanna Walton, Clément Crepy, Alicia Parrish, Zongwei Zhou, Clement Farabet, Carey Radebaugh, Praveen Srinivasan, Claudia van~der Salm, Andreas Fidjeland, Salvatore Scellato, Eri Latorre-Chimoto, Hanna Klimczak-Plucińska, David Bridson, Dario de~Cesare, Tom Hudson, Piermaria Mendolicchio, Lexi Walker, Alex Morris, Matthew Mauger, Alexey Guseynov, Alison Reid, Seth Odoom, Lucia Loher, Victor Cotruta, Madhavi Yenugula, Dominik Grewe, Anastasia Petrushkina, Tom Duerig, Antonio Sanchez, Steve Yadlowsky, Amy Shen, Amir Globerson, Lynette Webb, Sahil Dua, Dong Li, Surya Bhupatiraju, Dan Hurt, Haroon Qureshi, Ananth Agarwal, Tomer Shani, Matan Eyal, Anuj Khare, Shreyas~Rammohan Belle, Lei Wang, Chetan Tekur, Mihir~Sanjay Kale, Jinliang Wei, Ruoxin Sang, Brennan Saeta, Tyler Liechty, Yi~Sun, Yao Zhao, Stephan Lee, Pandu Nayak, Doug Fritz, Manish~Reddy Vuyyuru, John Aslanides, Nidhi Vyas,
  Martin Wicke, Xiao Ma, Evgenii Eltyshev, Nina Martin, Hardie Cate, James Manyika, Keyvan Amiri, Yelin Kim, Xi~Xiong, Kai Kang, Florian Luisier, Nilesh Tripuraneni, David Madras, Mandy Guo, Austin Waters, Oliver Wang, Joshua Ainslie, Jason Baldridge, Han Zhang, Garima Pruthi, Jakob Bauer, Feng Yang, Riham Mansour, Jason Gelman, Yang Xu, George Polovets, Ji~Liu, Honglong Cai, Warren Chen, XiangHai Sheng, Emily Xue, Sherjil Ozair, Christof Angermueller, Xiaowei Li, Anoop Sinha, Weiren Wang, Julia Wiesinger, Emmanouil Koukoumidis, Yuan Tian, Anand Iyer, Madhu Gurumurthy, Mark Goldenson, Parashar Shah, MK~Blake, Hongkun Yu, Anthony Urbanowicz, Jennimaria Palomaki, Chrisantha Fernando, Ken Durden, Harsh Mehta, Nikola Momchev, Elahe Rahimtoroghi, Maria Georgaki, Amit Raul, Sebastian Ruder, Morgan Redshaw, Jinhyuk Lee, Denny Zhou, Komal Jalan, Dinghua Li, Blake Hechtman, Parker Schuh, Milad Nasr, Kieran Milan, Vladimir Mikulik, Juliana Franco, Tim Green, Nam Nguyen, Joe Kelley, Aroma Mahendru, Andrea Hu, Joshua
  Howland, Ben Vargas, Jeffrey Hui, Kshitij Bansal, Vikram Rao, Rakesh Ghiya, Emma Wang, Ke~Ye, Jean~Michel Sarr, Melanie~Moranski Preston, Madeleine Elish, Steve Li, Aakash Kaku, Jigar Gupta, Ice Pasupat, Da-Cheng Juan, Milan Someswar, Tejvi M., Xinyun Chen, Aida Amini, Alex Fabrikant, Eric Chu, Xuanyi Dong, Amruta Muthal, Senaka Buthpitiya, Sarthak Jauhari, Nan Hua, Urvashi Khandelwal, Ayal Hitron, Jie Ren, Larissa Rinaldi, Shahar Drath, Avigail Dabush, Nan-Jiang Jiang, Harshal Godhia, Uli Sachs, Anthony Chen, Yicheng Fan, Hagai Taitelbaum, Hila Noga, Zhuyun Dai, James Wang, Chen Liang, Jenny Hamer, Chun-Sung Ferng, Chenel Elkind, Aviel Atias, Paulina Lee, Vít Listík, Mathias Carlen, Jan van~de Kerkhof, Marcin Pikus, Krunoslav Zaher, Paul Müller, Sasha Zykova, Richard Stefanec, Vitaly Gatsko, Christoph Hirnschall, Ashwin Sethi, Xingyu~Federico Xu, Chetan Ahuja, Beth Tsai, Anca Stefanoiu, Bo~Feng, Keshav Dhandhania, Manish Katyal, Akshay Gupta, Atharva Parulekar, Divya Pitta, Jing Zhao, Vivaan Bhatia,
  Yashodha Bhavnani, Omar Alhadlaq, Xiaolin Li, Peter Danenberg, Dennis Tu, Alex Pine, Vera Filippova, Abhipso Ghosh, Ben Limonchik, Bhargava Urala, Chaitanya~Krishna Lanka, Derik Clive, Yi~Sun, Edward Li, Hao Wu, Kevin Hongtongsak, Ianna Li, Kalind Thakkar, Kuanysh Omarov, Kushal Majmundar, Michael Alverson, Michael Kucharski, Mohak Patel, Mudit Jain, Maksim Zabelin, Paolo Pelagatti, Rohan Kohli, Saurabh Kumar, Joseph Kim, Swetha Sankar, Vineet Shah, Lakshmi Ramachandruni, Xiangkai Zeng, Ben Bariach, Laura Weidinger, Tu~Vu, Alek Andreev, Antoine He, Kevin Hui, Sheleem Kashem, Amar Subramanya, Sissie Hsiao, Demis Hassabis, Koray Kavukcuoglu, Adam Sadovsky, Quoc Le, Trevor Strohman, Yonghui Wu, Slav Petrov, Jeffrey Dean, and Oriol Vinyals. 2024.
\newblock \href {https://arxiv.org/abs/2312.11805} {Gemini: A family of highly capable multimodal models}.
\newblock \emph{Preprint}, arXiv:2312.11805.

\bibitem[{Thorne et~al.(2018)Thorne, Vlachos, Christodoulopoulos, and Mittal}]{ref_Fever}
James Thorne, Andreas Vlachos, Christos Christodoulopoulos, and Arpit Mittal. 2018.
\newblock \href {https://doi.org/10.18653/v1/N18-1074} {{FEVER}: a large-scale dataset for fact extraction and {VER}ification}.
\newblock In \emph{Proceedings of the 2018 Conference of the North {A}merican Chapter of the Association for Computational Linguistics: Human Language Technologies, Volume 1 (Long Papers)}, pages 809--819, New Orleans, Louisiana. Association for Computational Linguistics.

\bibitem[{Touvron et~al.(2023)Touvron, Lavril, Izacard, Martinet, Lachaux, Lacroix, Rozière, Goyal, Hambro, Azhar, Rodriguez, Joulin, Grave, and Lample}]{touvron2023llamaopenefficientfoundation}
Hugo Touvron, Thibaut Lavril, Gautier Izacard, Xavier Martinet, Marie-Anne Lachaux, Timothée Lacroix, Baptiste Rozière, Naman Goyal, Eric Hambro, Faisal Azhar, Aurelien Rodriguez, Armand Joulin, Edouard Grave, and Guillaume Lample. 2023.
\newblock \href {https://arxiv.org/abs/2302.13971} {Llama: Open and efficient foundation language models}.
\newblock \emph{Preprint}, arXiv:2302.13971.

\bibitem[{Tran et~al.(2024{\natexlab{a}})Tran, Khanh, Nguyen~Tuong, Dang, Nguyen, T.~Thinh, and T.~Hung}]{Tran_2024}
Bao Tran, T.~N. Khanh, Khang Nguyen~Tuong, Thien Dang, Quang Nguyen, Nguyen T.~Thinh, and Vo~T.~Hung. 2024{\natexlab{a}}.
\newblock \href {https://doi.org/10.1007/978-3-031-74127-2_19} {\emph{BERT-Based Model for Vietnamese Fact Verification Dataset}}, page 219–231.
\newblock Springer Nature Switzerland.

\bibitem[{Tran et~al.(2024{\natexlab{b}})Tran, Tran, and Tran}]{ref_ViNSV}
Quang-Duy Tran, Thai-Hoa Tran, and Khanh~Quoc Tran. 2024{\natexlab{b}}.
\newblock \href {https://doi.org/10.1109/MAPR63514.2024.10660734} {Advancing vietnamese fact extraction and verification through multi-stage text ranking}.
\newblock In \emph{2024 International Conference on Multimedia Analysis and Pattern Recognition (MAPR)}, pages 1--7.

\bibitem[{Wang et~al.(2021)Wang, Song, and Zhu}]{wang2021ensemblelearningbasedclassification}
Guangtao Wang, Qinbao Song, and Xiaoyan Zhu. 2021.
\newblock \href {https://arxiv.org/abs/2101.05993} {Ensemble learning based classification algorithm recommendation}.
\newblock \emph{Preprint}, arXiv:2101.05993.

\bibitem[{Yuan and Vlachos(2024)}]{ref_Zero-Shot}
Moy Yuan and Andreas Vlachos. 2024.
\newblock \href {https://doi.org/10.18653/v1/2024.kallm-1.11} {Zero-shot fact-checking with semantic triples and knowledge graphs}.
\newblock In \emph{Proceedings of the 1st Workshop on Knowledge Graphs and Large Language Models (KaLLM 2024)}, pages 105--115, Bangkok, Thailand. Association for Computational Linguistics.

\bibitem[{Zhong et~al.(2020)Zhong, Xu, Tang, Xu, Duan, Zhou, Wang, and Yin}]{ref_ROLGFC}
Wanjun Zhong, Jingjing Xu, Duyu Tang, Zenan Xu, Nan Duan, Ming Zhou, Jiahai Wang, and Jian Yin. 2020.
\newblock \href {https://doi.org/10.18653/v1/2020.acl-main.549} {Reasoning over semantic-level graph for fact checking}.
\newblock In \emph{Proceedings of the 58th Annual Meeting of the Association for Computational Linguistics}, pages 6170--6180, Online. Association for Computational Linguistics.

\bibitem[{Zhou et~al.(2019)Zhou, Han, Yang, Liu, Wang, Li, and Sun}]{ref_gear}
Jie Zhou, Xu~Han, Cheng Yang, Zhiyuan Liu, Lifeng Wang, Changcheng Li, and Maosong Sun. 2019.
\newblock \href {https://doi.org/10.18653/v1/P19-1085} {{GEAR}: Graph-based evidence aggregating and reasoning for fact verification}.
\newblock In \emph{Proceedings of the 57th Annual Meeting of the Association for Computational Linguistics}, pages 892--901, Florence, Italy. Association for Computational Linguistics.

\end{thebibliography}

\clearpage

\onecolumn

\appendix
\section{Comprehensive Data Processing} \label{appendix:data_processing}
\subsection{Dataset Statistics} \label{appendix:data-statistics}

This section presents the detailed statistics of the two datasets used in our experiments: ISE-DSC01 and ViWikiFC. The table below summarizes the number of samples in the training, development, and test sets for both corpora. These statistics help contextualize the scale and evaluation coverage of our methods.

\begin{table}[h]
  \centering
  \resizebox{0.3\columnwidth}{!}{%
    \begin{tabular}{lcc} 
      \hline
      & \textbf{ISE-DSC01} & \textbf{ViWikiFC} \\  
      \hline
      \textbf{Train} & 37,967 & 16,738 \\ 
      \textbf{Dev} & 4,794 & 2,090 \\ 
      \textbf{Test} & 5,396 & 2,091 \\ 
      \hline
    \end{tabular}%
  }
  \caption{Dataset statistics for ISE-DSC01 and ViWikiFC.}
  \label{tab:data-statistics}
\end{table}

\subsection{Data Analysis on Context Lengths}
\label{appendix:data_analysis}

\begin{figure}[h]
  \centering
  \includegraphics[width=0.8\columnwidth]{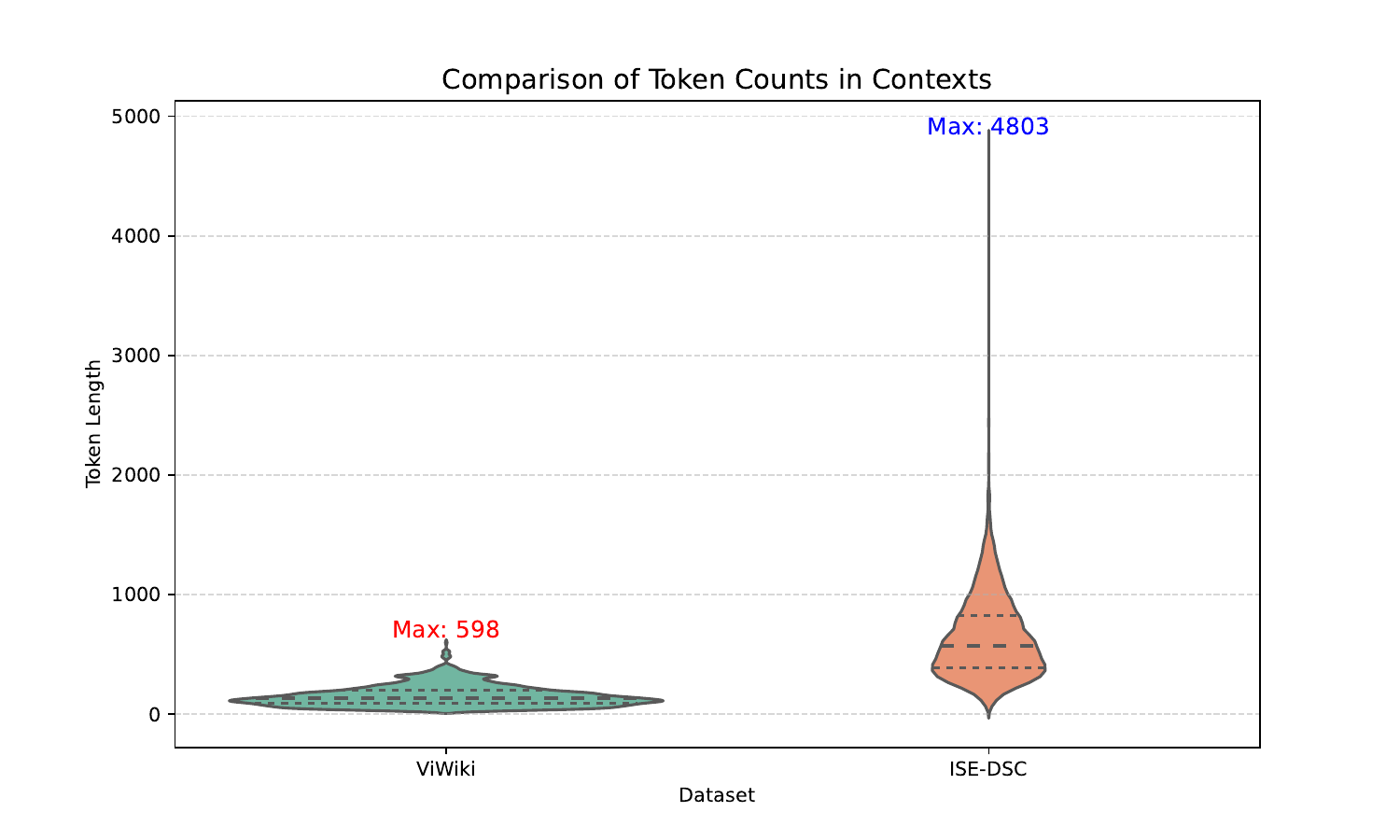}
  \caption{Graph representing the lengths of contexts.}
  \label{fig:data_analysis}
\end{figure}

Figure~\ref{fig:data_analysis} illustrates the distribution of token lengths across input contexts in the ViWikiFC and ISE-DSC01 datasets. As shown, many samples significantly exceed the 512-token input limitation of standard Transformer models. The ISE-DSC01 dataset, in particular, contains several contexts with over 4,000 tokens. This analysis highlights the necessity for effective context segmentation strategies to ensure full coverage of relevant evidence while maintaining compatibility with model constraints.

% \subsection{Long Context Processing Strategy}
\label{appendix:long_context}

Figure~\ref{fig:long_context} illustrates our strategy for handling long input contexts. First, the context is segmented into individual sentences. Next, sentences are sequentially aggregated into subcontexts until reaching approximately 400 tokens. Each completed subcontext is then passed to the QATC model to identify potential evidence. The next subcontext begins from the subsequent sentence, continuing until all sentences have been processed. However, processing subcontexts sequentially can be time-consuming. Therefore, we developed SER Faster, which batches subcontexts and processes them in parallel, significantly accelerating the retrieval process.

\section{Strict Accuracy in Fact-Checking}
\label{appendix:strict_acc}

 \textbf{Strict Accuracy}: This metric is a stringent measure that requires both the verdict and the evidence to be predicted correctly compared to the ground truth sample.

Verdict (\(v\) and \(v'\)): refers to the verdict of the sample and the predicted verdict (supported, refuted, nei).

Evidence (\(e\) and \(e'\)): refers to the evidence of the sample and the predicted evidence.

\begin{equation}
StrAcc = f(v,v').f(e,e')
 \end{equation}

Where:
\begin{equation}
f(v, v') = \begin{cases} 
1 & \text{if } v = v' \\
0 & \text{otherwise} 
\end{cases}
\end{equation}

\begin{equation}
f(e, e') = \begin{cases} 
1 & \text{if } e = e' \\
0 & \text{otherwise} 
\end{cases}
\end{equation}

Strict accuracy is the average of all StrAcc values.

\section{Ablation Study}
\label{appendix:Ablation}
\subsection{Ablation Experiments}
\begin{table}[h]
  \centering
  \resizebox{0.8\columnwidth}{!}{%
  \begin{tabular}{ll| ccc ccc}
    \hline
    \multicolumn{2}{c|}{Configuration} & \multicolumn{3}{c}{ViWikiFC} & \multicolumn{3}{c}{ISE-DSC01} \\
    \hline
    \textbf{ER Model} & \textbf{VC Model} & Strict Acc & VC Acc & ER Acc & Strict Acc & VC Acc & ER Acc \\
    \hline
    \multicolumn{8}{l}{\textbf{SemViQA (Full)}} \\
    \multirow{3}{*}{InfoXLM\(_{\text{large}}\)} & InfoXLM\(_{\text{large}}\) & 80.68 & \textbf{83.98} & \textbf{95.31} & 75.13 & 79.60 & 76.87 \\
    & XLM-R\(_{\text{large}}\) & \textbf{80.82} & 83.88 & \textbf{95.31} & 76.74 & 81.71 & 78.95 \\
    & Ernie-M\(_{\text{large}}\) & 80.06 & 83.17 & \textbf{95.31} & \textbf{78.97} & \textbf{82.49} & \textbf{80.91} \\
    \hline
    \multicolumn{8}{l}{\textbf{w/o Binary Classification (One-step VC)}} \\
    \multirow{3}{*}{InfoXLM\(_{\text{large}}\)} & InfoXLM\(_{\text{large}}\) & 79.63 & 82.88 & 95.31 & 73.87 & 78.35 & 76.89 \\
    & XLM-R\(_{\text{large}}\) & 80.73 & 83.69 & 95.31 & 75.96 & 80.80 & 78.97 \\
    & Ernie-M\(_{\text{large}}\) & 79.91 & 83.07 & 95.31 & 78.47 & 81.89 & 80.93 \\
    \hline
    \multicolumn{8}{l}{\textbf{w/o QATC (TF-IDF-based ER)}} \\
    \multirow{3}{*}{TF-IDF} & InfoXLM\(_{\text{large}}\) & 76.57 & 83.26 & 90.15 & 74.89 & 79.36 & 76.61 \\
    & XLM-R\(_{\text{large}}\) & 76.47 & 82.93 & 90.15 & 76.39 & 81.41 & 78.58 \\
    & Ernie-M\(_{\text{large}}\) & 75.75 & 81.97 & 90.15 & 78.71 & 82.28 & 80.65 \\
    \hline
  \end{tabular}
  }
  \caption{Ablation results of SemViQA.}
  \label{tab:ablation_results}
\end{table}

Table~\ref{tab:ablation_results} presents the ablation results to evaluate the contribution of each component in the SemViQA framework. When employing the full model with both the QATC-based evidence retrieval and the two-step verdict classification (TVC), SemViQA achieves the best performance across both datasets. Notably, using Ernie-M\(_{large}\) yields the highest strict accuracy (78.97\%) and evidence retrieval accuracy (80.91\%) on the ISE-DSC01 dataset. Removing the binary classification stage (i.e., using only one-step VC) leads to a noticeable performance drop, especially on ISE-DSC01, indicating that the binary classifier enhances the distinction between \textit{Supported} and \textit{Refuted} labels. Furthermore, replacing the QATC module with a TF-IDF based retriever results in a significant decline (about 5\%) in evidence retrieval accuracy, which subsequently affects the overall performance. These findings highlight the critical role of both QATC and the two-step classification scheme in improving SemViQA's effectiveness on fact verification in Vietnamese.
\subsection{Analysis of Confidence Threshold in SemViQA}

\begin{figure}[h]
  \centering
  \includegraphics[width=1\linewidth]{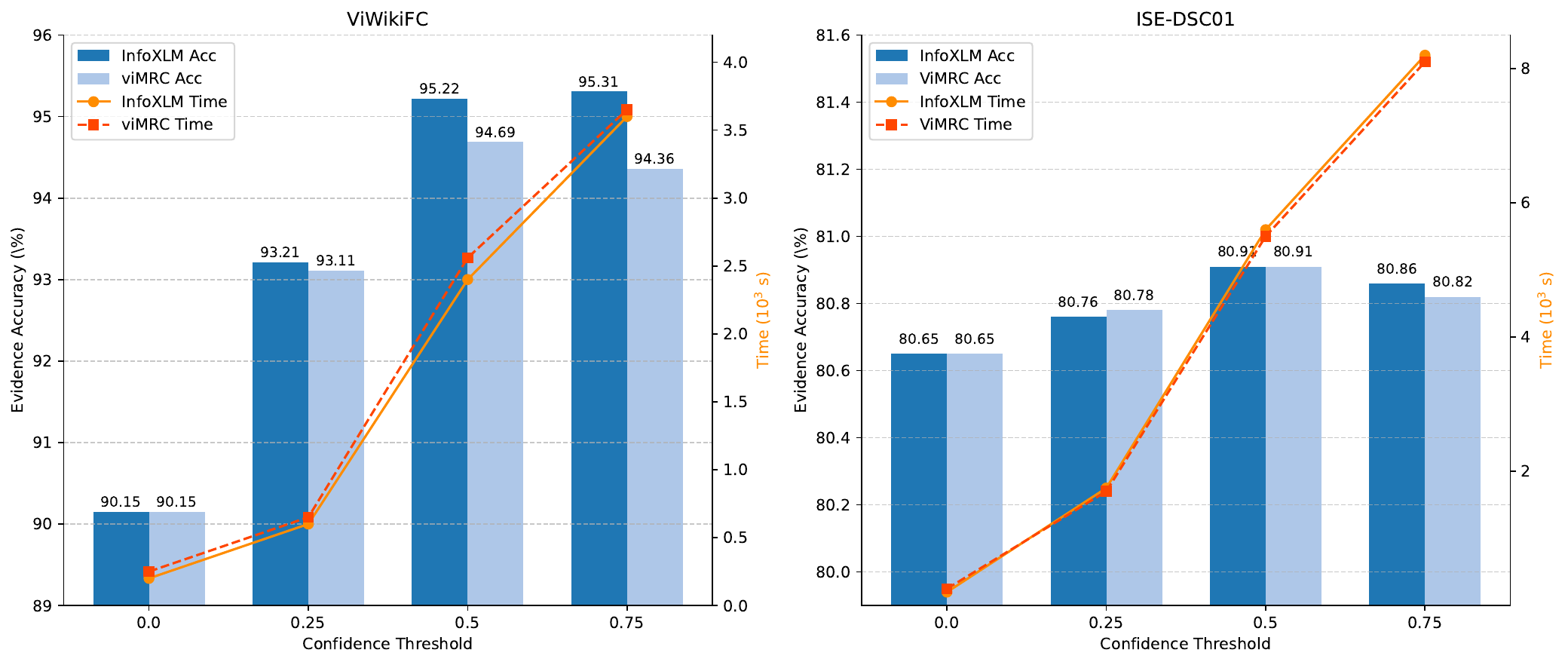}
  \caption{Impact of confidence threshold on evidence retrieval accuracy in SemViQA.}
  \label{fig:confidence_threshold}
\end{figure} 

The confidence threshold plays a crucial role in balancing accuracy and inference time in SemViQA's evidence retrieval process. Analysis from Figure \ref{fig:confidence_threshold} indicates that as the threshold increases from 0.0 to 0.5, evidence retrieval accuracy improves significantly, particularly on ViWikiFC (~95\%) and ISE-DSC01 (~80.8\%). However, beyond 0.5, accuracy gains plateau, while inference time decreases sharply due to the system filtering out low-confidence evidence more aggressively. Setting an optimal threshold in the range of 0.4 - 0.5 achieves a trade-off between efficiency and accuracy, ensuring that SemViQA operates swiftly while maintaining precise evidence retrieval.

\subsection{Effectiveness of QATC over Traditional QA-based Evidence Retrieval}

\begin{table}[h]
  \centering
  \resizebox{0.8\columnwidth}{!}{%
  \begin{tabular}{ll| ccc ccc}
    \hline
    \multicolumn{2}{c|}{SemViQA} & \multicolumn{3}{c}{ViWikiFC} & \multicolumn{3}{c}{ISE-DSC01} \\
    \hline
    SER & TVC & Strict Acc & VC Acc & ER Acc & Strict Acc & VC Acc & ER Acc \\
    \hline
    \multicolumn{8}{l}{\textbf{QATC-based ER}} \\
    \multirow{3}{*}{InfoXLM\(_{\text{large}}\)} & InfoXLM\(_{\text{large}}\) & 80.68 & \textbf{83.98} & \textbf{95.31} & 75.13 & 79.60 & 76.87 \\
    & XLM-R\(_{\text{large}}\) & \textbf{80.82} & 83.88 & \textbf{95.31} & 76.74 & 81.71 & 78.95 \\
    & Ernie-M\(_{\text{large}}\) & 80.06 & 83.17 & \textbf{95.31} & \textbf{78.97} & \textbf{82.49} & \textbf{80.91} \\
    \hline
    \multirow{3}{*}{ViMRC\(_{large}\)} & InfoXLM\(_{large}\) & 80.25 & 83.84 & 94.69 & 75.13 & 79.54 & 76.87\\
    & XLM-R\(_{large}\) & 80.34 & 83.64 & 94.69 & 76.71 & 81.65 & 78.91\\
    & Ernie-M\(_{large}\) & 79.53 & 82.97 & 94.69 & \textbf{78.97} & \textbf{82.54} & \textbf{80.91} \\
    \hline
    \multicolumn{8}{l}{\textbf{QA-based ER}} \\
    \multirow{3}{*}{InfoXLM\(_{\text{large}}\)} & InfoXLM\(_{\text{large}}\) & 79.96 & 83.50 & 94.45 & 74.02 & 78.95 & 75.83 \\
    & XLM-R\(_{\text{large}}\) & 80.11 & 83.60 & 94.45 & 75.61 & 80.95 & 77.91 \\
    & Ernie-M\(_{\text{large}}\) & 79.24 & 82.74 & 94.45 & 77.82 & 81.76 & 79.82 \\
    \hline
    \multirow{3}{*}{ViMRC\(_{large}\)} & InfoXLM\(_{large}\) & 79.77 & 83.84 & 94.26 & 74.05 & 78.93 & 75.87 \\
    & XLM-R\(_{\text{large}}\) & 79.87 & 83.79 & 94.26 & 75.65 & 80.93 & 77.95 \\
    & Ernie-M\(_{\text{large}}\) & 79.01 & 82.78 & 94.26 & 77.84 & 81.73 & 79.86 \\

    \hline
  \end{tabular}
  }
  \caption{Comparison of QATC-based vs. QA-based evidence retrieval in SemViQA. QATC consistently improves all metrics.}
  \label{tab:qatc_vs_qa_in_semviqa}
\end{table}

To compare the learning capabilities of QATC with traditional QA models, we construct two versions of the SemViQA pipeline that are identical in structure, differing only in the evidence retrieval model. As shown in Table~\ref{tab:qatc_vs_qa_in_semviqa}, using QATC consistently outperforms QA-based retrieval across both datasets. Specifically, on ViWikiFC, QATC achieves up to 80.82\% Strict Accuracy and 95.31\% ER Accuracy, while QA-based models peak at 80.11\% and 94.45\%, respectively. The improvement is even more evident on ISE-DSC01, where QATC reaches 78.97\% Strict Accuracy and 80.91\% ER Accuracy. These results confirm that QATC is more effective at learning to identify relevant evidence, leading to superior performance across the entire fact-checking pipeline.

\subsection{Analysis of Confusion Matrix in Verdict Classification}

\begin{figure}[h]
  \centering
  \includegraphics[width=1\linewidth]{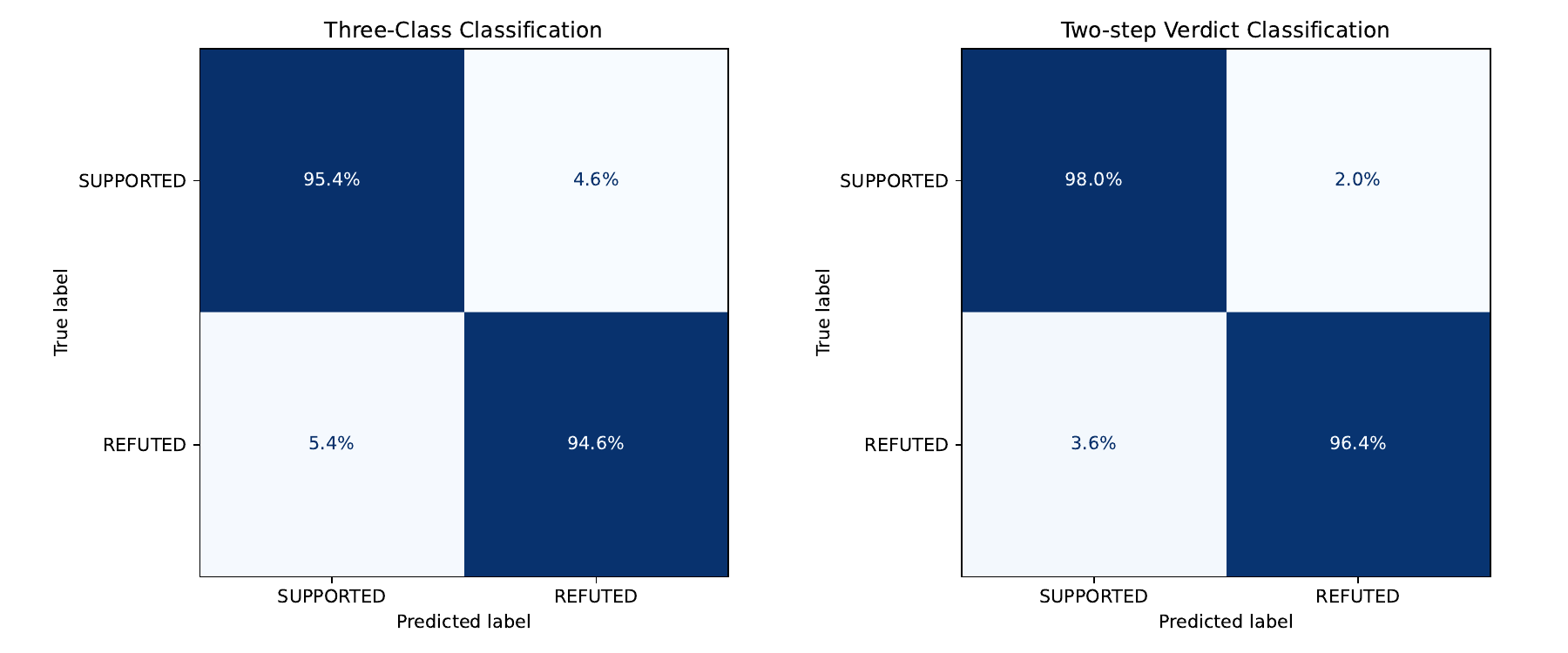}
  \caption{Confusion matrix of the Three-Class and Two-Step Verdict Classification (TVC) on the ISE-DSC01 dataset.}
  \label{fig:confusion_matrix_tc_tvc}
\end{figure} 

To evaluate the effectiveness of our Two-step Verdict Classification (TVC) strategy, we analyze the confusion matrices in Figure~\ref{fig:confusion_matrix_tc_tvc}, which compare the standard three-class classification with our proposed two-step approach, focusing solely on \textbf{Supported} and \textbf{Refuted} claims. The three-class classifier achieves strong performance, with accuracy of 95.4\% for Supported and 94.6\% for Refuted; however, it still exhibits notable confusion between the two classes, likely due to their semantic proximity and shared evidence patterns. In contrast, the two-step approach—by isolating NEI cases early and applying a dedicated binary classifier further improves accuracy to 98.0\% for Supported and 96.4\% for Refuted. These results support our hypothesis that decomposing the verdict classification task into semantically coherent subtasks enhances the model's precision in detecting factual consistency.

\section{Hyperparameter and LLM Training Configuration}
\label{appendix:hyperparameter_llm}

In this section, we present the detailed hyperparameter settings and training configurations for both our SemViQA models and the Large Language Model (LLM) fine-tuning process. Table~\ref{tab:hyperparameter_llm} consolidates all hyperparameters used across different models, including Binary Classification (BC), Three-Class Classification (TC), Question Answering with Token Classification (QATC), and LLM fine-tuning.

\begin{table}[h]
  \centering
  \resizebox{0.5\columnwidth}{!}{%
    \begin{tabular}{l|cccc}
      \hline
      \textbf{Hyperparameter} & \textbf{BC} & \textbf{TC} & \textbf{QATC} & \textbf{LLM} \\
      \hline
      Epochs & 20  & 20 & 20 & 1 \\
      RT Loss & -   & - & \checkmark & -  \\ 
      Cross-Entropy Loss & -   & \checkmark & \checkmark    & -\\
      Focal Loss & \checkmark & - & - & -    \\
      Learning Rate & $1e^{-5}$ & $1e^{-5}$ & $2e^{-6}$ & $5e^{-5}$ \\
      Batch Size & 104   & 104 & 36 & 2  \\
      Gradient Accumulation & 1  & 1 & 2 & 1  \\
      Optimizer (AdamW) & \checkmark & \checkmark & \checkmark & \checkmark  \\ 
      Max Token Length & 256 & 256 & 512 & 4096  \\ 
      GPUs  & A100 & A100 & A100 & A100  \\ 
      Zero & - & - & - & Zero3  \\ 
      LR Schedule  & Linear & Linear & Cyclic & Cosine  \\  
      Mixed Precision  & - & - & - & bf16  \\ 
      \hline
    \end{tabular}%
  }
  \caption{Consolidated hyperparameter and training configuration for SemViQA models and LLM fine-tuning.}
  \label{tab:hyperparameter_llm}
\end{table}

We fine-tune a Large Language Model (LLM) using a restructured version of the original datasets, ViWikiFC and ISE-DSC01, as detailed in Figure~\ref{fig:full_width_box_prompt_llm_train}. These datasets have been carefully adapted for training to improve performance and ensure compatibility with our model. For training, we utilize the official Qwen LLM implementation from the QwenLM repository\footnote{\url{https://github.com/QwenLM/Qwen}}. Our training setup follows the full configuration outlined in Table~\ref{tab:hyperparameter_llm}, ensuring optimal efficiency and alignment with best practices.

\clearpage
\begin{table*}[!htbp]
    \centering
    \begin{mdframed}[backgroundcolor=gray!5, linewidth=0.75pt, roundcorner=5pt]
    
    \textbf{Question:} You are tasked with verifying the correctness of the following statement. \\
    - We provide you with a claim and a context. Please classify the claim into one of three labels: ``Supported'', ``REFUTED'', or ``NEI'' (Not Enough Info).\\  
    - Your answer should include the classification label and the most relevant evidence sentence from the context.  \\
    - Remember, the evidence must be a full sentence, not part of a sentence or less than one sentence.

    \vspace{2mm}  % Giúp tách biệt các phần trong hộp để dễ đọc hơn
    Given a claim and context as follows:
    
    \textbf{Context:}  
    \small
    \textit{The actress revealed her secrets to maintaining a youthful appearance as follows: Eating three balanced meals a day. For dinner, Ivy Chen usually eats early to ensure her body has enough time to digest food, metabolize energy, and avoid putting pressure on the stomach and other organs. A recent study published in *Frontiers in Nutrition* suggests that eating dinner earlier can lead to a longer lifespan, with the ideal time being 7 PM. If this is not possible, experts recommend having the last meal of the day 2-3 hours before bedtime. Drinking ginger tea: To keep her body warm, promote blood circulation, and enhance circulation, Ivy Chen drinks ginger tea daily. Her ginger tea is typically made with ground ginger, black tea, turmeric powder, and brown sugar. This drink is a natural remedy that not only boosts the immune system and reduces inflammation but also fights oxidation, supports weight loss, improves skin health, and helps maintain a youthful look. Regular exercise: Ivy Chen is a fitness enthusiast who loves physical activities and exercises daily, even during pregnancy. The Taiwanese actress shared that if she is not busy with work, she runs for at least 30 minutes every day. \textcolor{blue}{Even when traveling abroad, she maintains her running habit.} A recent study published in *Progress in Cardiovascular Disease* found that regular runners live three years longer than non-runners. Running significantly helps with weight loss, maintaining a balanced physique, toning muscles, relaxing the mind, and benefiting heart health. Besides running, Ivy Chen also swims, practices yoga, and hikes to maintain physical fitness and endurance. Skincare: Regarding her skincare routine, the actress emphasized the importance of hydration. The Taiwanese beauty revealed that she always carries a facial mist to ensure her skin stays hydrated while outdoors.}
    
    \vspace{2mm}
    \textbf{Claim:}  
    \textit{Even when traveling abroad, Ivy Chen maintains her running habit.}
    
    \vspace{2mm}
    \textbf{Answer:}  
    This claim is classified as \textbf{Supported}.  
    The evidence is: \textit{\textcolor{blue}{Even when traveling abroad, she maintains her running habit.}}
    
    \end{mdframed}
    \caption{Example of a fact-checking task prompt used for LLM training. Note: Some parts of the Context and Claim were originally in Vietnamese. In this paper, we have translated them into English for better readability. Sentences highlighted in blue indicate the evidence.}
    \label{fig:full_width_box_prompt_llm_train}
\end{table*}

\begin{figure*}[h!]
  \centering
  \includegraphics[width=\linewidth]{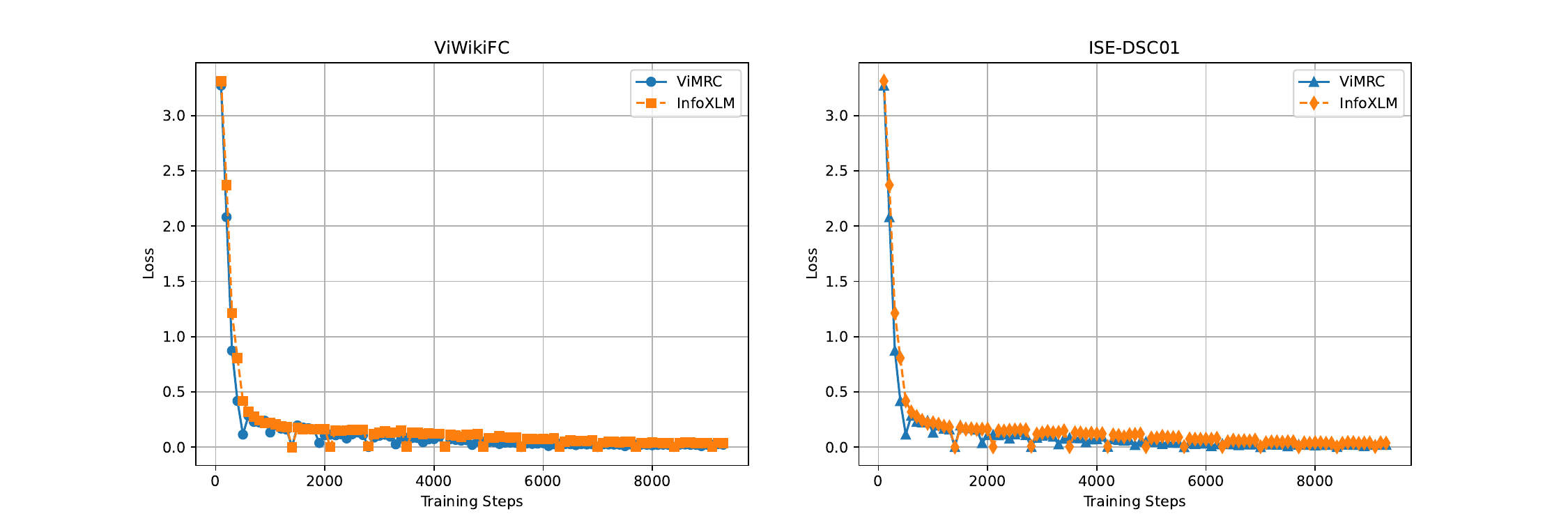}
  \caption{Training progress of the ViMRC\(_{\text{large}}\) and InfoXLM\(_{\text{large}}\) models.}
  \label{fig:loss_plot_QATC}
\end{figure*}

\begin{figure*}[h!]
  \centering
  \includegraphics[width=\linewidth]{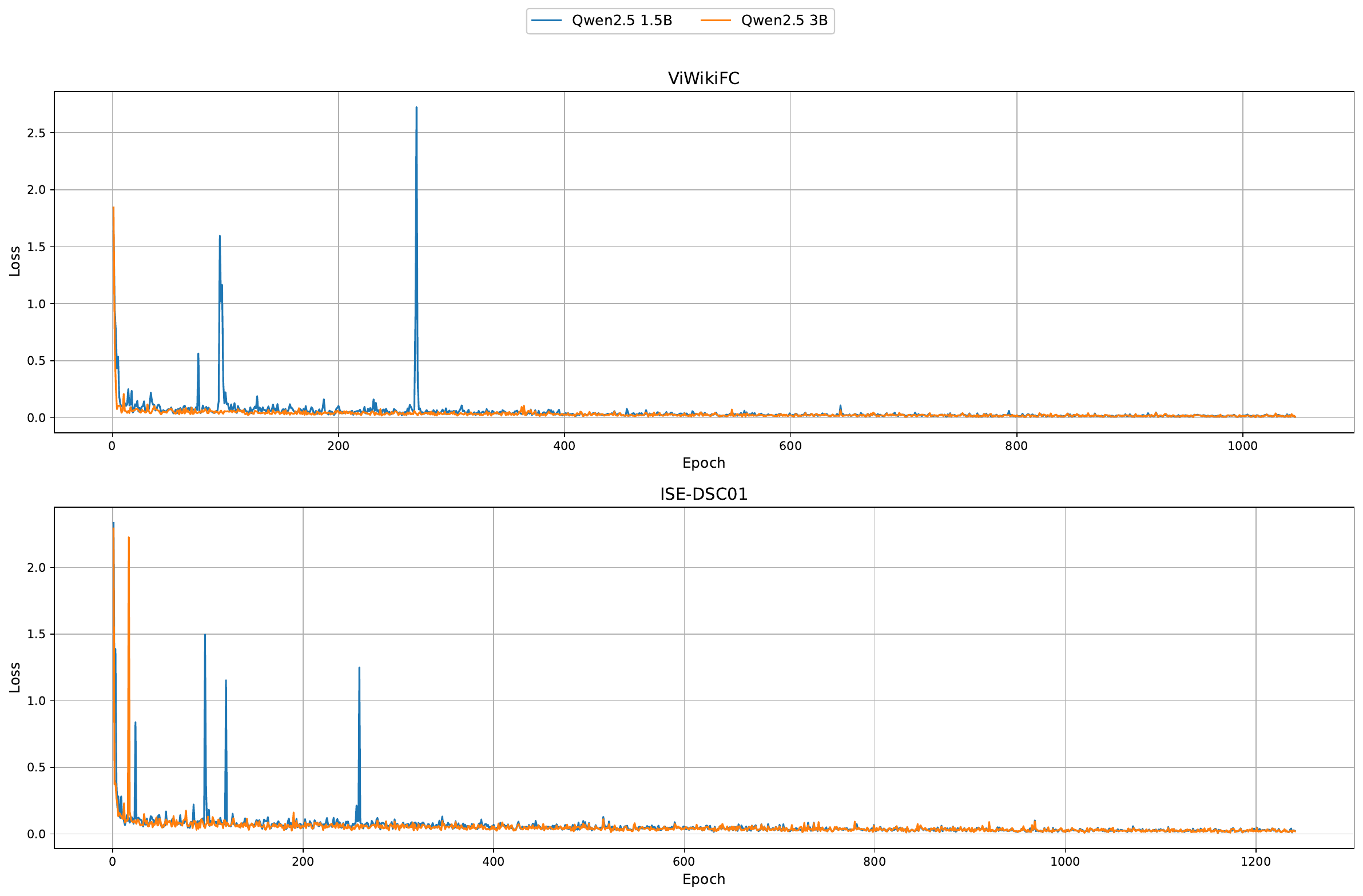}
  \caption{Training progress of the Qwen 1.5B and Qwen 3B models.}
  \label{fig:loss_plot_llm}
\end{figure*}

We present the complete training progress of the LLM models and QATC in Figure~\ref{fig:loss_plot_llm} and Figure~\ref{fig:loss_plot_QATC}, respectively. Figure~\ref{fig:loss_plot_llm} illustrates the training dynamics of Qwen 1.5B and Qwen 3B, supporting the results presented in Table~\ref{tab:main_results}. Notably, the Qwen 1.5B model demonstrates more stable training dynamics compared to the Qwen 3B model during the initial stage. Meanwhile, Figure~\ref{fig:loss_plot_QATC} showcases the completion of QATC training, depicting the loss curves of ViMRC\(_{\text{large}}\) and InfoXLM\(_{\text{large}}\). These results highlight the convergence behavior of QATC training across different architectures, further supporting the robustness of our approach.

\clearpage

\clearpage
\section{Comparison of TF-IDF and QATC in Fact-Checking: Examples of Incorrect vs. Correct Evidence Selection}
\label{appendix:tfidf_qatc_example}

\begin{table*}[htbp]
  \centering
  \begin{tabularx}{\textwidth}{|>{\raggedright\arraybackslash}X|>{\raggedright\arraybackslash}X|>{\raggedright\arraybackslash}X|>{\raggedright\arraybackslash}X|}
    \hline
    \textbf{Claim} & \textbf{Evidence} & \textbf{TF-IDF} & \textbf{QATC} \\  
    \hline
    Du lịch Triều Tiên là điều mà chỉ có một số người được đi đến.  
    \newline \textcolor{blue!50!white}{(Traveling to North Korea is something only a few people can do.)} &  
    Theo nguyên tắc, bất kỳ ai cũng được phép du lịch tới Triều Tiên, và những ai có thể hoàn thành quá trình làm thủ tục thì đều không bị Triều Tiên từ chối cho nhập cảnh.  
    \newline \textcolor{blue!50!white}{(In principle, anyone is allowed to travel to North Korea, and those who complete the process are not denied entry.)} &  
    Khách du lịch không được đi thăm thú bên ngoài vùng đã được cho phép trước mà không được hướng dẫn viên người Triều Tiên cho phép nhằm tránh các điệp viên nằm vùng.  
    \newline \textcolor{blue!50!white}{(Tourists are not allowed to visit areas outside of the designated zones without a North Korean guide to prevent undercover spies.)} &  
    Theo nguyên tắc, bất kỳ ai cũng được phép du lịch tới Triều Tiên, và những ai có thể hoàn thành quá trình làm thủ tục thì đều không bị Triều Tiên từ chối cho nhập cảnh.  
    \newline \textcolor{blue!50!white}{(In principle, anyone is allowed to travel to North Korea, and those who complete the process are not denied entry.)} \\
    \hline
    Nó có độ nóng chảy ở mức gần 30 độ C.  
    \newline \textcolor{blue!50!white}{(It has a melting point of about 30°C.)} &  
    Nó là một kim loại kiềm mềm, màu bạc, và với điểm nóng chảy là 28 °C (83 °F) khiến cho nó trở thành một trong các kim loại ở dạng lỏng tại hay gần nhiệt độ phòng.  
    \newline \textcolor{blue!50!white}{(It is a soft, silvery alkali metal with a melting point of 28°C (83°F), making it one of the metals that is liquid at or near room temperature.)} &  
    Nó là nguyên tố có độ âm điện thấp thứ hai sau franci, và chỉ có một đồng vị bền là caesi-133.  
    \newline \textcolor{blue!50!white}{(It is the second least electronegative element after francium, and has only one stable isotope, cesium-133.)} &  
    Nó là một kim loại kiềm mềm, màu bạc, và với điểm nóng chảy là 28 °C (83 °F) khiến cho nó trở thành một trong các kim loại ở dạng lỏng tại hay gần nhiệt độ phòng.  
    \newline \textcolor{blue!50!white}{(It is a soft, silvery alkali metal with a melting point of 28°C (83°F), making it one of the metals that is liquid at or near room temperature.)} \\
    \hline
  \end{tabularx}
  \caption{Comparison of TF-IDF and QATC in Fact-Checking: TF-IDF selects irrelevant evidence (Incorrect), while QATC selects accurate evidence (Correct).}
  \label{tab:fact_checking_comparison_tfidf_vs_QATC}
\end{table*}

\end{document}